\documentclass[10pt,twocolumn,letterpaper]{article}

\usepackage[pagenumbers]{iccv} 

\definecolor{iccvblue}{rgb}{0.21,0.49,0.74}
\usepackage[pagebackref,breaklinks,colorlinks,allcolors=iccvblue]{hyperref}
\usepackage{multirow}

\usepackage[linesnumbered,ruled,vlined]{algorithm2e}
\usepackage[accsupp]{axessibility}  

\def\confName{ICCV}
\def\confYear{2025}

\newcommand{\ourmethod}{\textit{FaceLift}}

\title{\ourmethod: Learning Generalizable Single Image 3D Face Reconstruction from Synthetic Heads}

\author{Weijie Lyu$^{1 *}$ \quad Yi Zhou$^2$ \quad Ming-Hsuan Yang$^1$ \quad Zhixin Shu$^{2 \dag}$\vspace{0.2cm}\\
$^1$University of California, Merced \quad $^2$Adobe Research\vspace{0.2cm}\\
}
\newcommand\blfootnote[1]{%
  \begingroup
  \renewcommand\thefootnote{}\footnote{#1}%
  \addtocounter{footnote}{-1}%
  \endgroup
}

\begin{document}

\twocolumn[{%
\renewcommand\twocolumn[1][]{#1}%
\maketitle 
\begin{center} 
\vspace{-8mm}
\centering 
\includegraphics[width=0.97\linewidth]{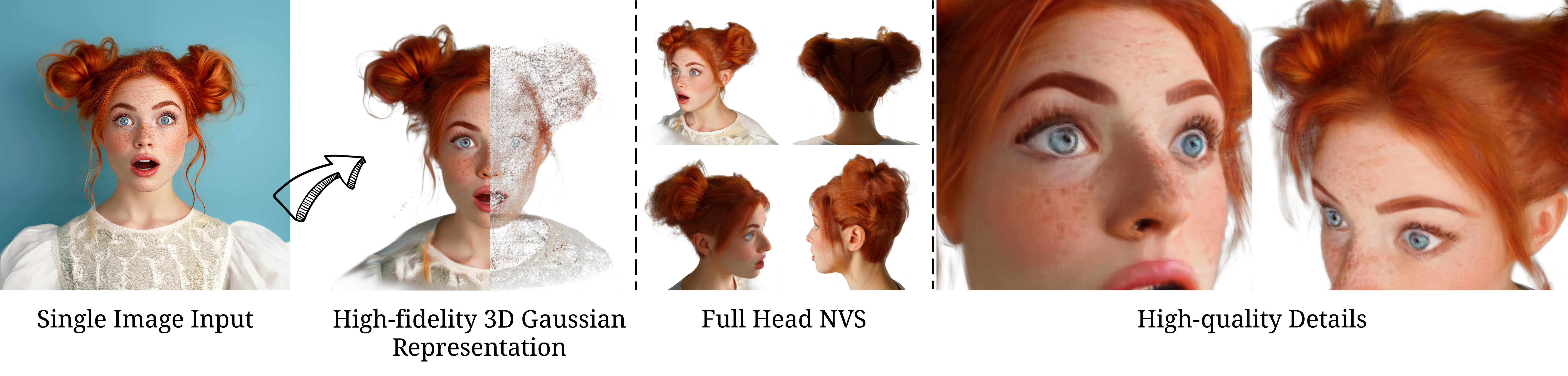}
    \vspace{-2mm}
    \captionof{figure}
    {{\ourmethod} transforms a single facial image into a high-fidelity 3D Gaussian head representation. Trained exclusively on synthetic 3D data, our pipeline first generates sparse, identity-preserving multiview images of the input head using a diffusion model. These sparse generated views are then fed into a transformer-based 3D Gaussian splats reconstructor, producing complete and detailed 3D head representation that generalize remarkably well to real-world human images. Project page: \url{https://weijielyu.github.io/FaceLift}.}
    \label{fig:teaser}
\end{center}
}]

\blfootnote{*Work was done when Weijie Lyu was an intern at Adobe Research.}
\blfootnote{\dag Corresponding author.}

\vspace{-4mm}
\begin{abstract}
We present {\ourmethod}, a novel feed-forward approach for generalizable high-quality 360-degree 3D head reconstruction from a single image.
Our pipeline first employs a multi-view latent diffusion model to generate consistent side and back views from a single facial input, which then feed into a transformer-based reconstructor that produces a comprehensive 3D Gaussian splats representation.
Previous methods for monocular 3D face reconstruction often lack full view coverage or view consistency due to insufficient multi-view supervision. We address this by creating a high-quality synthetic head dataset that enables consistent supervision across viewpoints.
%
To bridge the domain gap between synthetic training data and real-world images, we propose a simple yet effective technique that ensures the view generation process maintains fidelity to the input by learning to reconstruct the input image alongside the view generation.
%
%
Despite being trained exclusively on synthetic data, our method demonstrates remarkable generalization to real-world images.
Through extensive qualitative and quantitative evaluations, we show that {\ourmethod} outperforms state-of-the-art 3D face reconstruction methods on identity preservation, detail recovery and rendering quality.
%
\end{abstract}    
\section{Introduction}
\vspace{-1mm}
\label{sec:intro}
%
3D face reconstruction has been a central focus in computer vision and graphics research for decades, driven by its crucial applications in immersive virtual and augmented realities, VFX and gaming, digital entertainment, and next-generation telepresence systems.
%
However, achieving high quality reconstruction from a single image remains very challenging. On one hand, the monocular face reconstruction problem is highly ill-posed -- a single 2D image can be produced by countless different 3D face shapes, creating fundamental ambiguity. On the other hand, the human visual system is highly attuned to facial details, making even subtle artifacts and imperfections noticeable to the eye.

\begin{figure}[t]
    \centering
    \includegraphics[width=.47\textwidth]{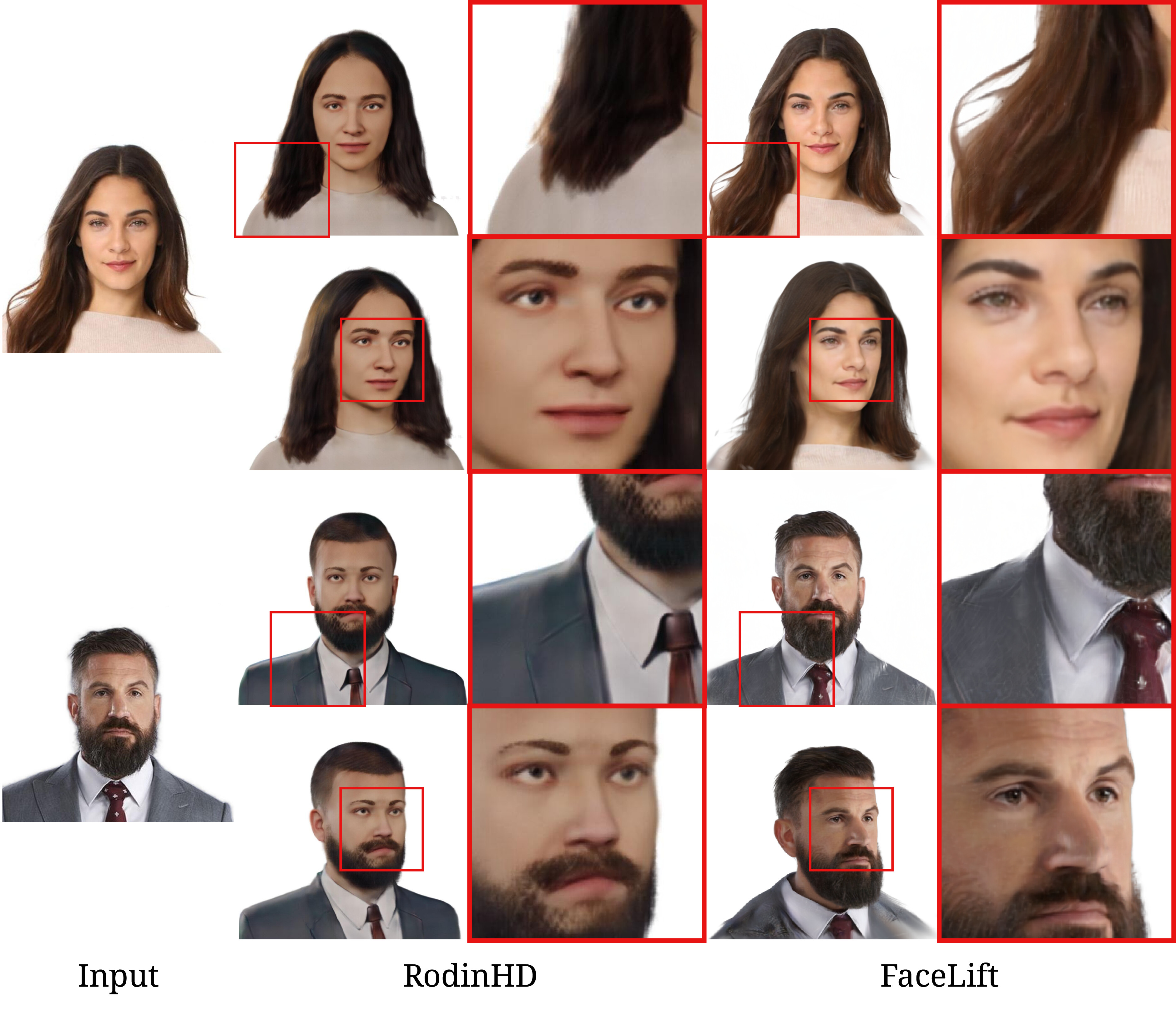}
    \vspace{-4mm}
    \caption{\textbf{Comparison with RodinHD.} RodinHD~\cite{rodinhd} trains triplane diffusion with synthetic data, resulting in apparent identity loss. In contrast, {\ourmethod} achieves better identity preservation and generalizes effectively to real human portraits.}
    \vspace{-6mm}
    \label{fig:rodinhd}
\end{figure}

Traditional 3D head synthesis approaches typically use parametric textured mesh models~\cite{vetter1998estimating, flame} trained on 3D scan datasets. While these models enable basic head generation, the rendered images frequently lack fine-scale geometric detail, realistic textures, and convincing hair, limiting their perceptual realism and expressiveness.
%
Recent breakthroughs in image generative models~\cite{gan, diffusion} and novel view synthesis techniques~\cite{nerf, gs} have opened new possibilities for this research area. Leveraging these developments, recent works~\cite{eg3d, panohead} use neural 3D representations to learn effective 3D head representation from unstructured real face image datasets~\cite{stylegan, lpff}.
While these datasets improve the realism and diversity of rendering results, they fail to provide multi-view supervision for modeling 3D consistency causing view inconsistency and artifacts on the back of the head.
Since diverse multi-view real images are hard to acquire, RodinHD~\cite{rodinhd} leverages synthetic multi-view images to train generative models that directly output 3D neural representations of the head. However, training solely on synthetic data often results in significant perceptual identity loss in the generated outputs, as demonstrated in Fig.~\ref{fig:rodinhd}.

In this work, we present {\ourmethod}, a pipeline for learning generalizable and high-fidelity single image to 3D face representation from synthetic head data.
%
To achieve high quality reconstruction that preserves the input facial identity, we adopt a two-stage pipeline to first generate identity preserving multi-view images using a diffusion model~\cite{latent_diffusion}, followed by a transformer-based feed-forward reconstructor to fuse the generated sparse views into a comprehensive 3D Gaussian representation. 
We train the model with synthetic images -- multi-view renderings of 3D synthetic human portraits using Blender. We highlight two key techniques for generalizing to real-world images and preserving input facial identity. First, we emphasize the importance of reconstructing the input image alongside the view synthesis task in the conditional diffusion model training, which significantly improved generalization capability in testing. Second, we demonstrate that training the feed-forward reconstructor benefits from a two-stage training process: pre-training on general objects~\cite{objaverse} to acquire a rich geometry and texture prior, followed by fine-tuning on synthetic human head data to capture head-specific geometry. With our two-stage approach, we focus on learning identity preservation in the image space during the first stage, achieving higher input fidelity compared to existing methods.


Comparing with prior art, we achieve three key advancements: (1) robust view consistency through multi-view attention and supervision, (2) improved generalization from training techniques and foundational model, ensuring accurate identity preservation, (3) high-quality facial texture and hair details via pixel-aligned Gaussian representation.

We extensively evaluate {\ourmethod} quantitatively and qualitatively across diverse datasets. Using real multi-view studio captures~\cite{ava256} and an independent synthetic human dataset~\cite{cafca}, our approach consistently surpasses previous state-of-the-art methods across all evaluation metrics. 
Through extensive testing on in-the-wild portrait images, we demonstrate that our method reconstructs complete 3D heads with fine-grained details, accurate identity preservation, and high visual fidelity. Comparisons and ablation studies confirm that multi-view consistent training data is crucial for high-fidelity face reconstruction.
Our contributions are summarized as follows:
\begin{itemize}
    \item We propose {\ourmethod}, a framework that reconstructs a high-fidelity 3D head from a single image using view generation and feed-forward reconstructor.
    

    \item Despite being trained solely on synthetic human head data, our approach shows no domain gap on real-world images, highlighting both the effectiveness of synthetic data and our model's robust generalization capabilities.

    \item We construct two benchmarks on single-image to 3D full head reconstruction tasks using the publicly available datasets Cafca~\cite{cafca} and Ava-256~\cite{ava256} to quantitatively evaluate models' performance on both reconstruction accuracy and identity preservation ability.
    
    \item Our comprehensive evaluation confirms that our approach achieves state-of-the-art performance in reconstruction accuracy and identity preservation.
\end{itemize}

\vspace{-1mm}
\section{Related Work}
\label{sec:related_work}
\vspace{-1mm}
\noindent\textbf{Face Reconstruction.}
3D face reconstruction has been a long-standing challenge in computer vision, with substantial progress driven by diverse approaches. Vetter and Blanz~\cite{vetter1998estimating} pioneer a method for synthesizing 3D faces by linearly blending multiple 3D templates, now widely known as blendshapes. This work establishes the foundation for 3D Morphable Models (3DMMs), which represent 3D face shapes and textures as principal components derived from scanned data. Subsequent research~\cite{morphable, booth20163d, flame, lin2020towards, 3d_face_recon} extend this framework, enabling the generation of new 3D faces by manipulating blending coefficients. 
However, these methods produce mesh-based representations that lack fine details and are limited to modeling the front of the face, excluding hair and 360-degree synthesis.
While 3DMM-based methods have been foundational, recent advances in deep learning, especially Generative Adversarial Networks (GANs)~\cite{gan, stylegan, stylegan2}, have greatly improved 3D face synthesis quality. EG3D~\cite{eg3d} uses a tri-plane NeRF representation with a pose-conditioned StyleGAN2~\cite{stylegan2} framework. Follow-up works~\cite{oneshotavatar, triplanenet} achieve single-image-to-3D generation through GAN inversion~\cite{Inverting}. Despite their success, these methods can only synthesize near-frontal views. To overcome this, PanoHead~\cite{panohead} introduces a tri-grid neural volume representation, enabling full 360-degree head synthesis. Unfortunately, it does not provide a 3D head representation for consistent multi-view rendering.
Recent efforts explore alternative representations for 3D face reconstruction from sparse input, such as a single image~\cite{pnrf, relightify, teglo} or few-shot images~\cite{preface}. However, these methods still require pre-instance optimization. Rodin~\cite{rodin} and its extension RodinHD~\cite{rodinhd} employ an image-conditioned diffusion model to generate a triplane representation of a human head for full-head novel view synthesis. Nevertheless, their triplane diffusion model is limited to synthetic data and struggles to achieve high-fidelity reconstructions from real-world images. For animatable 3D head avatars generations, Morphable Diffusion~\cite{morphable_diffusion} generates multi-view consistent images from a single image using a morphable mesh, while HeadGAP~\cite{headgap} generates 3D animatable head avatars using few-shot input, leveraging 3D head priors learned from large-scale data. In contrast, our work focuses on leveraging synthetic training data to produce high-fidelity, detailed 3D Gaussian head models.

\vspace{1mm}
\noindent\textbf{Synthetic Human Data.}
Capturing high-quality 3D data of real humans requires a controlled studio environment and costly photography equipment~\cite{ava256}. As an alternative, large-scale synthetic 3D head datasets have emerged as an effective and resource-efficient solution for tasks like human head reconstruction~\cite{fake, rodin, rodinhd, cafca} and photorealistic relighting~\cite{lumos, synthlight}, offering a scalable way to train models without the restrictions of real-world data acquisition. Inspired by these previous works, we aim to use synthetic data to improve the model's understanding of the human head and minimize the generalization gap between synthetic data training and real-world inference.

\vspace{1mm}
\noindent\textbf{Image or Text to 3D.}
Generative models have achieved remarkable success in 2D image generation with VAEs~\cite{aevb, ndrl}, GANs~\cite{gan, stylegan, stylegan2}, and diffusion models~\cite{diffusion, latent_diffusion, ddim}. Building on this success, extensive research has extended these models to 3D content generation~\cite{3dgan, get3d, convmesh, stylesdf}. Starting with DreamFusion~\cite{dreamfusion}, numerous works~\cite{zero123, magic123, dreamgaussian, prolificdreamer, zeronvs} try to distill NeRF~\cite{nerf, eg3d, neus} or 3D Gaussians~\cite{gs} representation from 2D image diffusion using a Score Distillation Sampling (SDS) loss.
These methods can produce high-quality results but often encounter challenges such as slow optimization, over-saturated colors, and the Janus problem.
To overcome these challenges, recent works~\cite{one2345, zero123plus, wonder3d, era3d, instant3d} generate multi-view images with high consistency, which can be directly used for 3D reconstruction with neural reconstruction methods~\cite{nerf, gs, neus}. However, optimizing NeRF or NeuS remains far from real-time performance.
Recent advancements in large 3D reconstruction models (LRMs)~\cite{lrm, instant3d, gslrm, lgm} offer a pathway to faster 3D reconstruction. Leveraging scalable transformer architectures~\cite{transformer, vit} and large datasets like Objaverse~\cite{objaverse, objaversexl}, these models effectively capture generalizable 3D priors. Unlike traditional pre-scene optimization methods~\cite{nerf, gs, neus}, LRMs employ a feed-forward approach, enabling the prediction of high-quality NeRF, mesh, or 3D Gaussian representations from sparse images in under a second.
However, most of these research efforts are applied to general objects, with limited or suboptimal results for 3D head reconstruction.
\begin{figure*}[ht]
    \centering
    \includegraphics[width=.95\linewidth]{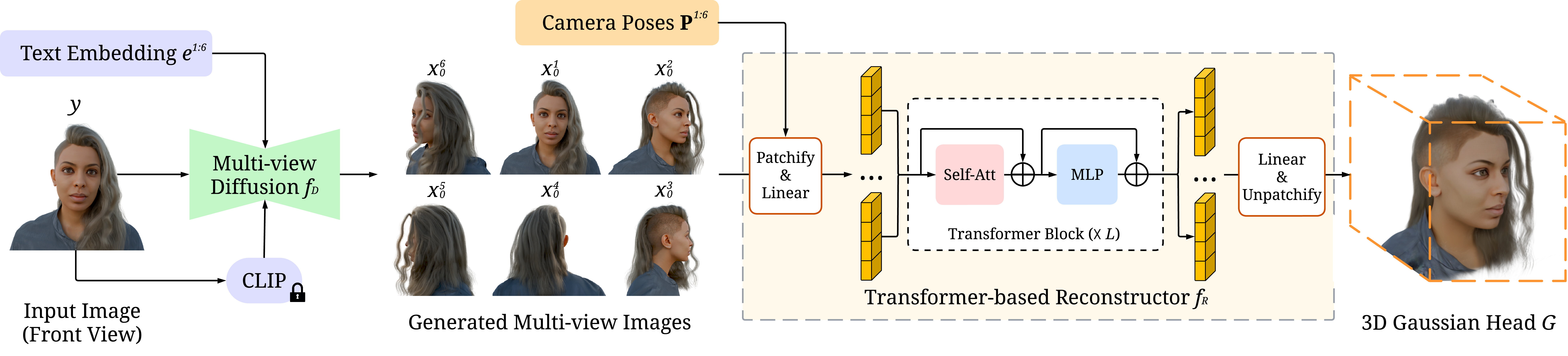}
    \vspace{-4mm}
    \caption{\textbf{Overview of {\ourmethod}.} Given a single image of a human face $y$ as input, we train an image-conditioned, multi-view diffusion model to generate novel views $x_0^1, \dots, x_0^N$ covering the entire head. By generating $x_0^1$ the same as $y$ and leveraging the high-quality synthetic data, our multi-view latent diffusion model can hallucinate unseen views of the human head with high-fidelity and multi-view consistency. We then train a transformer-based reconstructor $f_R$, which takes multi-view images $x_0^{1:N}$ and their camera poses $\textsc{P}^{1:N}$ as input and generates 3D Gaussian splats $G$ to represent the human head. 
    }
    \label{fig:method}
    \vspace{-4mm}
\end{figure*}
\vspace{-1mm}
\section{Proposed Method}
\vspace{-1mm}
\label{sec:proposed_method}
As shown in Fig.~\ref{fig:method}, given a single frontal image of a human face $y$, our goal is to reconstruct a complete 3D head $G$, represented as 3D Gaussian splats, with detailed facial texture and preserved identity.
This requires our system to have prior knowledge on the geometry structure of a human face and the ability to synthesis plausible details which are not visible in the input view.
%
%
Hence, we train a multi-view diffusion model $f_D$ on synthetic human head data to generate $N$ views $x_0^1, x_0^2, \dots, x_0^N$ covering $360^{\circ}$ of the human head while achieving multi-view consistency and preserving identity. 
We choose pixel-aligned 3D Gaussians to obtain the final 3D representation.
Compared to NeRFs and meshes, 3D Gaussians offer explicit volumetric primitives that better capture subtle facial geometry and fine details. Their semi-transparent kernels naturally model effects like wispy hair and translucency, which are challenging for discrete surfaces or density fields.
These generated views $x_0^{1:N}$ from the diffusion model, along with their corresponding Plücker ray coordinates $\textsc{P}^{(1:N)}$, are input into a transformer-based Gaussian reconstructor $f_R$ to predict a set of 3D Gaussians$G$. Training of the Gaussian reconstructor follows a pre-training process on general objects~\cite{objaverse} and a fine-tuning process on synthetic human head data.

\vspace{-1mm}
\subsection{Synthetic Human Head Dataset}
\vspace{-1mm}
\begin{figure}[!ht]
    \centering
    \includegraphics[width=.47\textwidth]{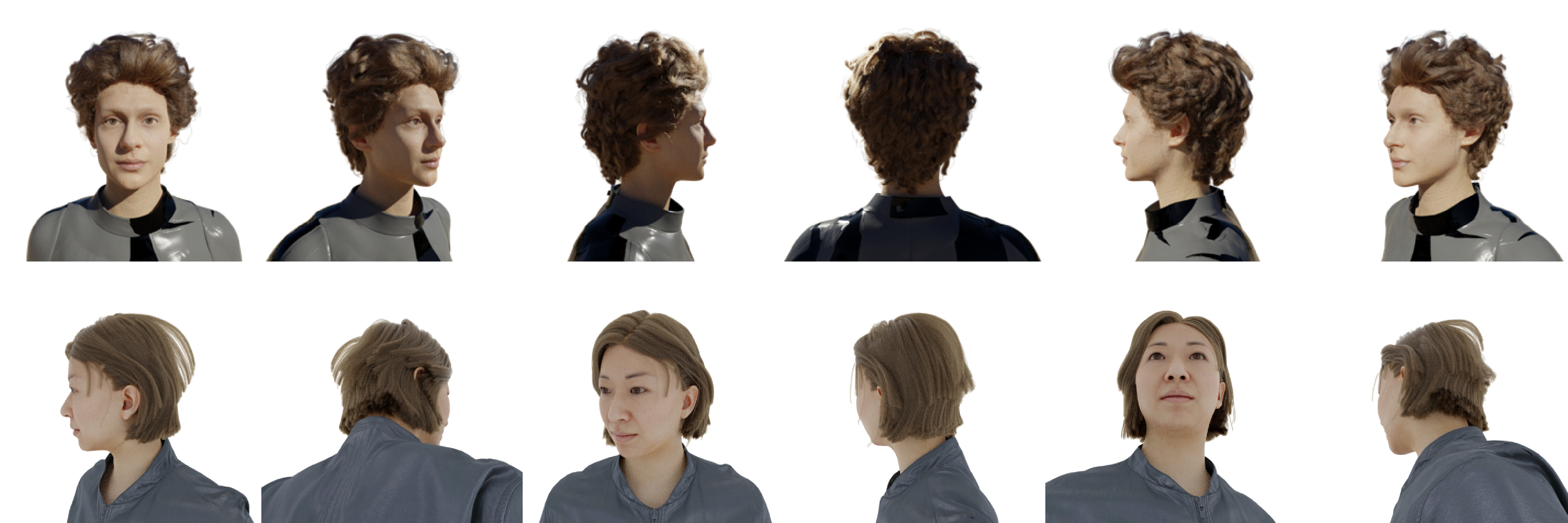}
    \vspace{-2mm}
    \caption{\textbf{Synthetic data examples.} Top row: six views for diffusion training. Bottom row: samples of random views for reconstructor training.}
    \label{fig:synthetic_data}
    \vspace{-4mm}
\end{figure}
We implement a 3D head asset generation pipeline inspired by~\cite{fake}. Our process begins with a collection of high-quality, artist-created 3D head meshes, which we enhance by incorporating detailed facial components, including eyes, teeth, gums, and both facial and scalp hair. We then augment these base models through rigging for pose variation and blendshape deformation for diverse facial expressions. The final head models are enriched with a set of PBR texture maps, including albedo, normal, roughness, specular, and subsurface scattering maps. At last, we dress the head model with a collection of clothing assets. The entire pipeline is implemented in Blender and the images are rendered with Cycles renderer.

To train our networks, we render images (samples shown in Fig.~\ref{fig:synthetic_data}) at 512$\times$512 resolution from 200 unique identities, each with 50 varied appearances, including different hairstyles, skin tones, expressions, clothes, poses, \etc. We render our training dataset under two types of lighting conditions: (1) ambient light and (2) random HDR environment light. We render six views for each subject to train the multi-view diffusion model. For fine-tuning the transformer-based reconstructor, we render 32 views with random camera poses.
\vspace{-1mm}
\subsection{View Generation}
\vspace{-1mm}
\label{subsec:mvdiffusion}
We model the sparse view generation from a single image input as a conditional diffusion process. We use a multi-view diffusion model \(f_D\) to generate $N$ views, denoted as $X_0^1, X_0^2, \dots, X_0^N$, given a single front-facing image $y$ and CLIP text embeddings $e^1, e^2, \dots, e^N$ corresponding to each generated view. This process is expressed as:
\vspace{-2mm}
\begin{equation}
    \{X_0^1, X_0^2, \dots, X_0^N\} = f_D(y, \{e^1, e^2, \dots, e^N\}).
\end{equation}
We aim to learn the joint distribution of all these views, conditioning on the input image $y$ and text embedding $e^1, e^2, \dots, e^N$. We denote the joint distribution as:
\vspace{-2mm}
\begin{equation}
    p_{f_D}(x_0^{1:N} \mid y, e^{1:N}) := p_{f_D}(\{x_0^1, \dots x_0^N\} \mid y, \{e^1, \dots e^N\}).
\end{equation}
\noindent\textbf{View Selection.}
Given a single near frontal view face image with azimuth $\alpha$, our multi-view diffusion model generates six views with azimuths equal to \{$\alpha$, $\alpha \pm 45^{\circ}$, $\alpha \pm 90^{\circ}$, $\alpha + 180^{\circ}$\}, covering 360 degrees of the human head. 
Elevation is 0 for all images.
We opt for six views as the optimal balance - fewer views compromise detail quality while more views become computationally prohibitive for full head reconstruction.
An ablation study comparing four, six, and eight views is presented in Sec.~\ref{subsec:num_of_views}.

\noindent\textbf{Multi-view Attention.}
To ensure the consistency of the generated novel views, we use a multi-view attention mechanism to facilitate information propagation and implicitly encode multi-view dependencies. 
%
%
Our attention module encourages multi-view consistency by extending the 2D self-attention mechanism to 3D and enabling interactions across views. Instead of treating each view independently, we apply self-attention across all views simultaneously, allowing information to be shared between them. Specifically, we start with an input tensor of shape $B \!\times\! V \!\times\! H \!\times\! W \!\times\! C$, where $B$ is the batch size, $V$ is the number of views, $H$ and $W$ denote the spatial resolution of the intermediate feature maps, and $C$ is the number of feature channels. We reshape this tensor to $B \!\times\! VHW \!\times\! C$, treating all spatial locations across all views as a unified sequence of tokens for self-attention. This design allows the model to learn multi-view correlations by sharing information across views within the attention layers, enabling it to generate consistent RGB images. We provide an ablation study on the multi-view attention mechanism in the supplementary material.

\noindent\textbf{Input View Reconstruction.}
%
%
During Training, we enforce the first generated view to share the same camera with the input image. In other words, we reconstruct the input view in the view generation process. We find this approach, combined with the multi-view attention mechanism, significantly outperforms the alternative strategy of generating only novel views, which tends to overfit to synthetic training identities and compromises generalization capability as we will show in Sec.~\ref{subsec:input_view_reconstruction} and the supplementary material.

\vspace{-1mm}
\subsection{Multi-view to 3D Gaussians Reconstruction}
\vspace{-1mm}
\label{subsec:gs-lrm}
\noindent\textbf{Transformer-based Reconstructor.}
We choose pixel-aligned 3D Gaussians as the final 3D representation. Each Gaussian $G_i$ is defined by position $p_i$, scale $s_i$, orientation $q_i$, opacity $\alpha_i$, and color features $c_i$. We use a transformer-based reconstructor $f_G$ to obtain 3D Gaussians from generated multi-view images $x_0^{1:N}$ and their corresponding Plücker ray coordinates~\cite{plucker1865xvii} \( \textsc{P}^{1:N} \):
\vspace{-2mm}
\begin{equation}
    \{G_i\}_{i=1}^{NHW}\!=\!\{p_i, s_i, q_i, \alpha_i, c_i\}_{i=1}^{NHW}\!=\!f_G(x_0^{1:N}, \textsc{P}^{1:N}),
\end{equation}
Our $f_G$ is a large reconstruction model~\cite{lrm, gslrm} which follows the implementation of GS-LRM~\cite{gslrm}: the $N$ multi-view images are concatenated with their Plücker ray coordinates computed from the camera intrinsic and extrinsic parameters for pose conditioning. Then, the inputs are patchified by dividing the per-view feature map into non-overlapping patches with a patch size of $p$. Each 2D patch is then flattened into a 1D vector. Finally, a linear layer $\textit{L}$ is utilized to map the 1D vectors to image patch tokens:
\vspace{-2mm}
\begin{equation}
\{\textsc{T}_{j}^n\}_{j=1,2,\dots,\frac{HW}{p^2}} = \textit{L}(\texttt{Patchify}_p(\texttt{Concat}(\textsc{I}^n, \textsc{P}^n))).
\end{equation}
Where $\{\textsc{T}_{j}^n\}$ denotes the set of patch tokens for image $n$, totaling $\frac{HW}{p^2}$ tokens per image.
The set of multi-view image tokens $\{\textsc{T}_j^n\}$ are concatenated and processed through a chain of transformer blocks. Each transformer block is equipped with residual connections~\cite{resnet} and consists of Pre-LayerNorm~\cite{layernorm}, multi-head Self-Attention~\cite{transformer} and MLP. Later, the output tokens from the transformer are decoded into Gaussian parameters using a single linear layer. Then, the Gaussian parameters are unpatchified into $p^2$ Gaussians. Finally, we end up with $HW$ Gaussians for each view, where pixel encodes one 3D Gaussian.

\vspace{1mm}
\noindent\textbf{Two-stage Training.}
We find that training the transformer-based reconstructor solely on synthetic human head data leads to inferior texture details when applied to real-world images (see ablation study in Fig.~\ref{fig:lrm_training}). We suspect this limitation arises because the synthetic datasets lack geometric diversity. To address this, we propose a two-stage training approach in which the reconstructor is pre-trained on diverse object data~\cite{objaverse} and subsequently fine-tuned using synthetic head data. The pre-training stage enables the reconstructor to learn a broad prior of diverse geometric structures, yielding more detailed and clearer textures in delicate facial regions such as the eyes, nose, and ears. The fine-tuning process then imparts specific knowledge of head structure, producing smoother and more realistic results.
%
%
During training, we randomly select four input views to reconstruct a total of eight views, four input and four novel views. Following~\cite{gslrm}, we optimize the model using a combination of MSE and perceptual losses. During inference, the reconstructor processes the six-view outputs from multi-view diffusion model to reconstruct the head.

\vspace{-1mm}
\subsection{Real-world Image Inference}
\vspace{-1mm}
\label{subsec:inference}
For inference on real-world images, since their intrinsic parameters are unknown, we adopt a camera fov of $50^\circ$, same as during training. To ensure plausible outputs, we first apply an MTCNN face detector to estimate the face's size and center. The image is then resized and cropped/extended to match the average face size and center computed from the training data. We find this alignment compensates for the unknown intrinsic parameters well, ensuring plausible reconstruction results.

\section{Experiments}
\label{sec:experiments}
\vspace{-1mm}
\subsection{Experimental Setup}
\vspace{-1mm}
\noindent\textbf{Evaluation Datasets.}
To quantitatively evaluate {\ourmethod}, we establish two benchmarks for single-image to 3D full head reconstruction tasks using publicly available datasets: (1) \textit{Cafca dataset}~\cite{cafca}: We select 40 subjects with 7 to 19 test camera poses each.
Since the camera positions are randomly distributed, we manually select the most frontal view as input.
Note that this synthetic dataset was independently developed and differs significantly from our training dataset.
(2) \textit{Ava-256 dataset}~\cite{ava256}: This studio-captured dataset contains real human subjects. We sample 10 subjects and 10 test camera poses for our evaluation. More details in supplemental.
To demonstrate our system's generalization capabilities, we also evaluate on a set of in-the-wild face images for qualitative assessment.

\vspace{1mm}
\noindent\textbf{Baselines.}
We compare our results against three state-of-the-art methods for single-face 3D reconstruction: GGHead~\cite{gghead}, PanoHead~\cite{panohead}, and Dual Encoder~\cite{dualencoder}. We perform GAN-inversion to reconstruct 3D head from a given input image using these models.
%
%
To emphasize the importance of utilizing our synthetic human head data for training, we also compare our method with two methods that focus on general object reconstruction: Era3D~\cite{era3d} and LGM~\cite{lgm}. More comparison results with mesh-based methods are provided in the supplementary material.

%
We further developed a baseline, \textit{Our MV + LGM}, which leverages the multi-view outputs generated by our diffusion model and employs LGM for reconstruction. This demonstrates that our method can be seamlessly integrated with other reconstruction frameworks to enhance performance on face reconstruction tasks. We tried to fine-tune the LGM reconstructor with our synthetic data, but it provides inferior results with incorrect geometry and artifacts compared with the original weights, which we suspect is due to training data mismatch (see details in the supplementary material).

\vspace{1mm}
\noindent\textbf{Evaluation Metrics.}
We evaluate reconstruction quality using four standard metrics: PSNR, SSIM, LPIPS~\cite{lpips}, and DreamSim~\cite{dreamsim}. To evaluate identity preservation, we perform face verification using ArcFace~\cite{arcface} through the DeepFace~\cite{deepface} implementation.

\vspace{1mm}
\noindent\textbf{Implementation Details.}
Both Cafca~\cite{cafca} and Ava-256~\cite{ava256} datasets provide multi-view RGB images and corresponding camera poses. However, their camera system differs from the ones utilized in {\ourmethod} and baselines.
We recalculate the test camera extrinsic in each method's camera system.
For a more accurate comparison, we use the Mediapipe facial landmark detection algorithm~\cite{mediapipe} to identify facial landmarks in both the target images and the rendered outputs, aligning them based on landmark distributions. Details are provided in the supplementary material. 

Our system takes approximately 8 seconds to infer a 3D Gaussian head from a single image: about 1.5 seconds for preprocessing (background removal, rescaling, etc.), 5.5 seconds for multi-view image generation, and under 1 second for 3D Gaussians reconstruction.

\begin{table}[tb]
  \centering
  \scriptsize
  \begin{tabular}{@{\hskip 2.0mm}l@{\hskip 2.0mm}c@{\hskip 2.0mm}c@{\hskip 2.0mm}c@{\hskip 2.0mm}c@{\hskip 2.0mm}c@{\hskip 2.0mm}}
    \toprule
    Method & PSNR $\uparrow$ & SSIM $\uparrow$ & LPIPS $\downarrow$ & DreamSim $\downarrow$ & ArcFace $\downarrow$ \\
    \midrule
    GGHead~\cite{gghead} & 10.35 & 0.7406 & 0.3636 & 0.3252 & 0.2681 \\
    PanoHead~\cite{panohead} & 10.72 & 0.7594 & 0.3351 & 0.2048 & 0.2183 \\
    Dual Encoder~\cite{dualencoder} & 10.78 & 0.7385 & 0.3922 & 0.2785 & 0.2421 \\
    Era3D~\cite{era3d} & 13.69 & 0.7230 & 0.3662 & 0.2892 & 0.2978 \\
    LGM~\cite{lgm} & 16.52 & 0.7933 & 0.3060 & 0.1552 & 0.2557 \\
    Our MV + LGM~\cite{lgm} & 14.13 & 0.7812 & 0.2956 & 0.1282 & 0.1767 \\
    \ourmethod & \textbf{16.61} & \textbf{0.7968} & \textbf{0.2694} & \textbf{0.1096} & \textbf{0.1573} \\
  \bottomrule
  \end{tabular}
  \vspace{-2mm}
    \caption{\textbf{Quantitative results on Cafca.} 
    {\ourmethod} achieves favorable performance on all evaluation metrics. 
    }
  \label{tab:cafca}
  \vspace{-2mm}
\end{table}

\begin{figure}[tb]
    \centering
    \includegraphics[width=.47\textwidth]{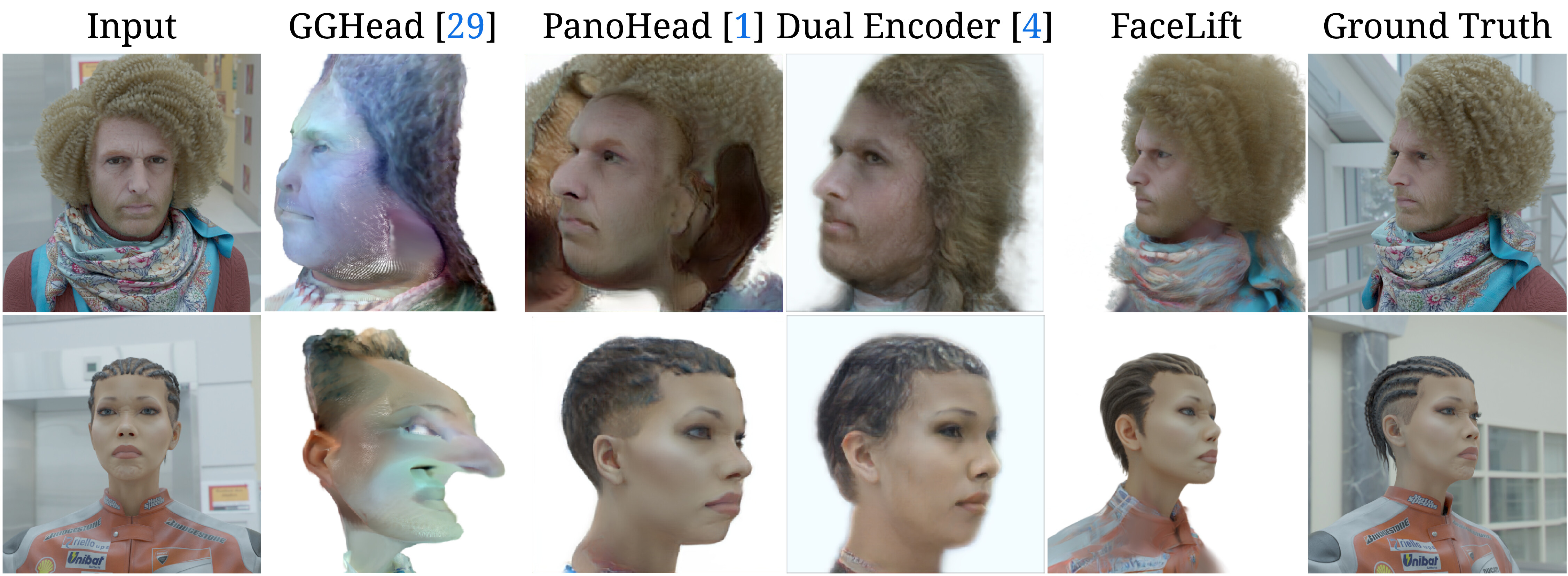}
    \vspace{-2mm}
    \caption{\textbf{Visual results on Cafca compared with face reconstruction methods.} {\ourmethod} renders novel views that closely match the ground truth, while other baselines often fail to reconstruct the 3D head in correct colors or geometry structures.}
    \label{fig:cafca}
    \vspace{-2mm}
\end{figure}
\begin{figure}[tb]
    \centering
    \includegraphics[width=.47\textwidth]{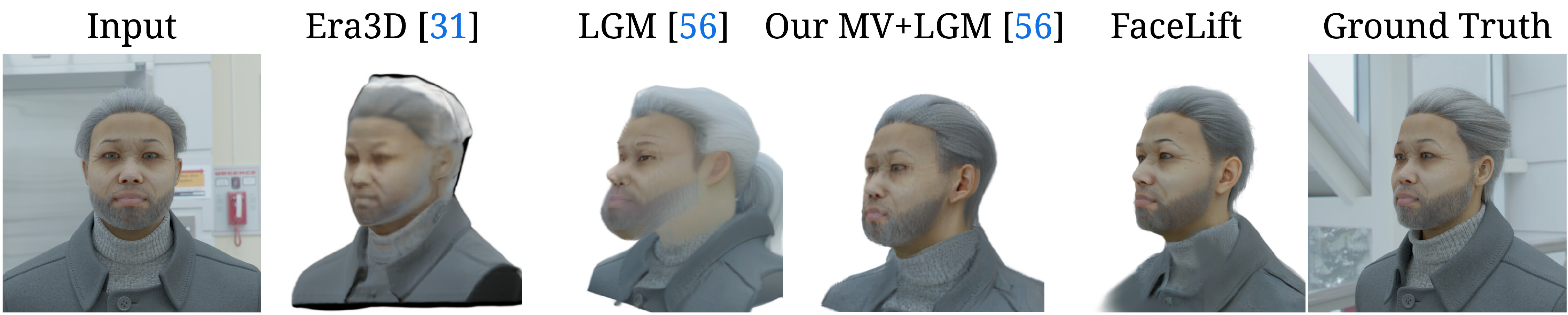}
    \vspace{-2mm}
    \caption{\textbf{Visual results on Cafca compared with general objects reconstruction methods.} Comparison with general object reconstruction methods shows the importance of specialized data.}
    \label{fig:cafca_lgm}
    \vspace{-4mm}
\end{figure}

\vspace{-1mm}
\subsection{Experiments on the Cafca Dataset}
\vspace{-1mm}
We report numerical comparison results on Cafca in Tab.~\ref{tab:cafca}. {\ourmethod} performs favorably against baselines, especially on DreamSim~\cite{dreamsim} metric, indicating high-quality perceptual similarity. It also achieves better identity preservation performance, as demonstrated by a lower ArcFace~\cite{arcface} distance. We show visual results in Fig.~\ref{fig:cafca} and Fig.~\ref{fig:cafca_lgm}. 
%
{\ourmethod} yields rendering results that closely match the ground truth. Compared with other baselines, GGHead~\cite{gghead} does not support full-head rendering, resulting in unrealistic outputs when the view angle significantly deviates from the input. PanoHead~\cite{panohead} struggles with challenging hairstyles, while Dual Encoder~\cite{dualencoder} produces blurred facial textures. Additionally, Era3D~\cite{era3d} introduces artifacts on the back of the head, and LGM~\cite{lgm} yields inaccurate nose and jaw shapes, underscoring the importance of our synthetic human head data. When integrated with our multi-view diffusion approach, LGM achieves enhanced performance, demonstrating that our method can be seamlessly combined with existing baselines to boost their results.

\begin{table}[tb]
  \centering
  \scriptsize
  \begin{tabular}{@{\hskip 2.0mm}l@{\hskip 2.0mm}c@{\hskip 2.0mm}c@{\hskip 2.0mm}c@{\hskip 2.0mm}c@{\hskip 2.0mm}c@{\hskip 2.0mm}}
    \toprule
    Method & PSNR $\uparrow$ & SSIM $\uparrow$ & LPIPS $\downarrow$ & DreamSim $\downarrow$ & ArcFace $\downarrow$ \\
    \midrule
    Era3D~\cite{era3d} & 14.77 & 0.7963 & 0.2538 & 0.2515 & 0.3721 \\
    LGM~\cite{lgm} & 14.05 & 0.8136 & 0.2476 & 0.1496 & 0.3142 \\
    Our MV+LGM~\cite{lgm} & 15.24 & 0.8213 & 0.2292 & 0.1093 & 0.2264 \\
    \ourmethod & \textbf{16.52} & \textbf{0.8271} & \textbf{0.2277} & \textbf{0.1065} & \textbf{0.1871} \\
  \bottomrule
  \end{tabular}
  \vspace{-2mm}
  \caption{\textbf{Quantitative results on Ava-256.} {\ourmethod} performs favorably than baseline methods in both reconstruction metrics and identity facial identity metric, showing a better generalization ability towards real-captured human images.}
  \label{tab:ava_256}
  \vspace{-2mm}
\end{table}

\begin{figure}[t]
    \centering
    \includegraphics[width=.47\textwidth]{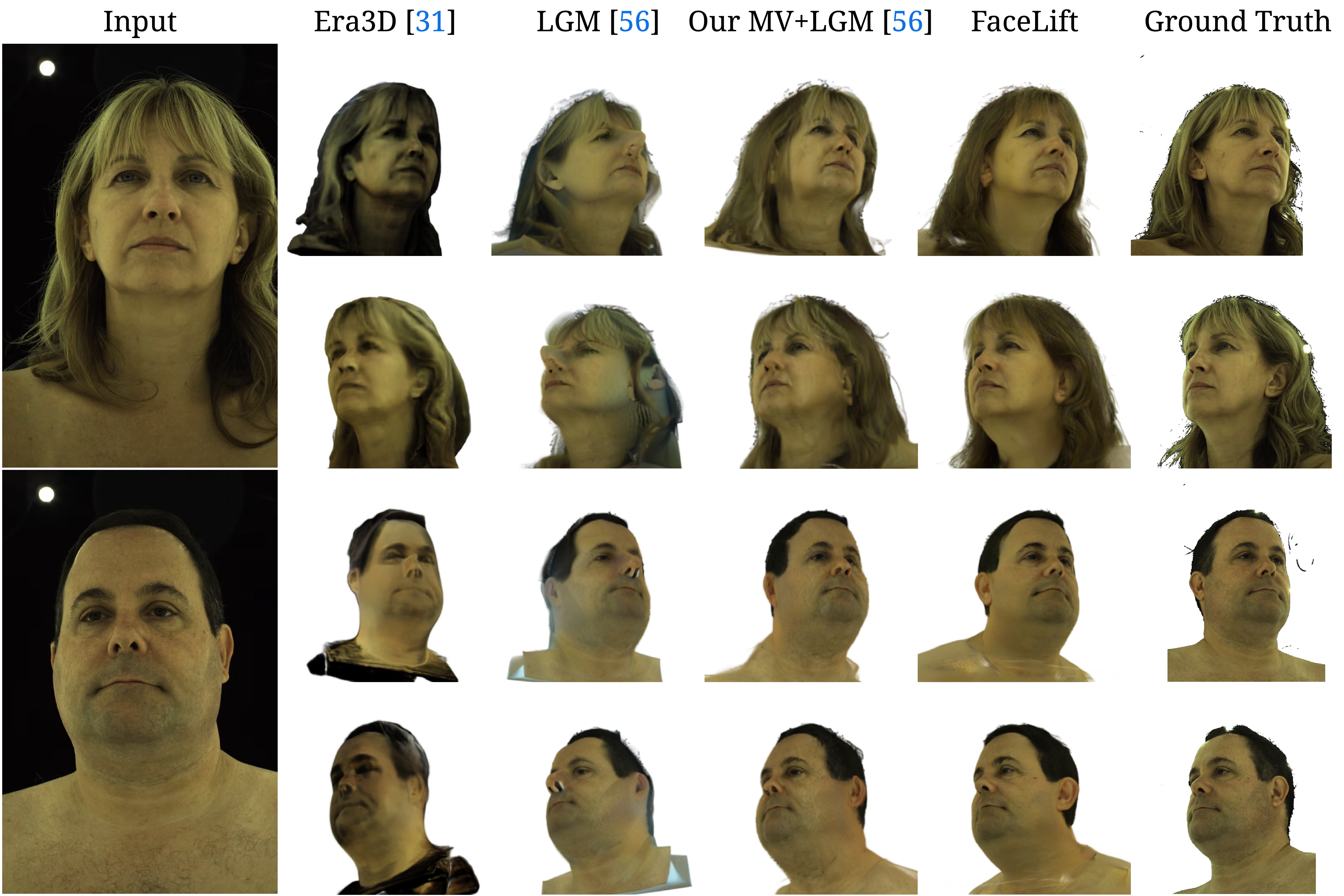}
    \vspace{-2mm}
    \caption{\textbf{Visual results on Ava-256.} Compared with baselines, {\ourmethod} provides multi-view renderings that are more realistic and similar to ground truth. Era3D fails to deliver delicate facial structures, while LGM generates inaccurate head shapes and colors.}
    \label{fig:ava_256}
    \vspace{-4mm}
\end{figure}
\vspace{-1mm}
\subsection{Experiments on the Ava-256 Dataset}
\vspace{-1mm}
We further evaluate {\ourmethod} against other baselines on a studio-captured real human dataset, Ava-256~\cite{rgca}. GAN-inversion based methods~\cite{panohead, gghead, dualencoder} fail to produce reasonable results with the test camera poses in this dataset, so we exclude these baselines. Tab.~\ref{tab:ava_256} shows that {\ourmethod} outperforms all other baselines across all evaluation metrics, demonstrating superior reconstruction quality and identity preservation. It also highlights {\ourmethod}'s strong ability to generalize to real human faces. As shown in Fig.~\ref{fig:ava_256}, {\ourmethod} achieves more realistic head synthesis, while Era3D~\cite{era3d} struggles with accurate skin and hair textures, as well as facial details. LGM~\cite{lgm} produces inaccuracies in the nose shape. When combined with our multi-view diffusion model, LGM yields more accurate geometric structures, yet its texture quality remains inferior to that of {\ourmethod}.

\begin{figure*}[tb]
    \centering
    \includegraphics[width=.95\textwidth]{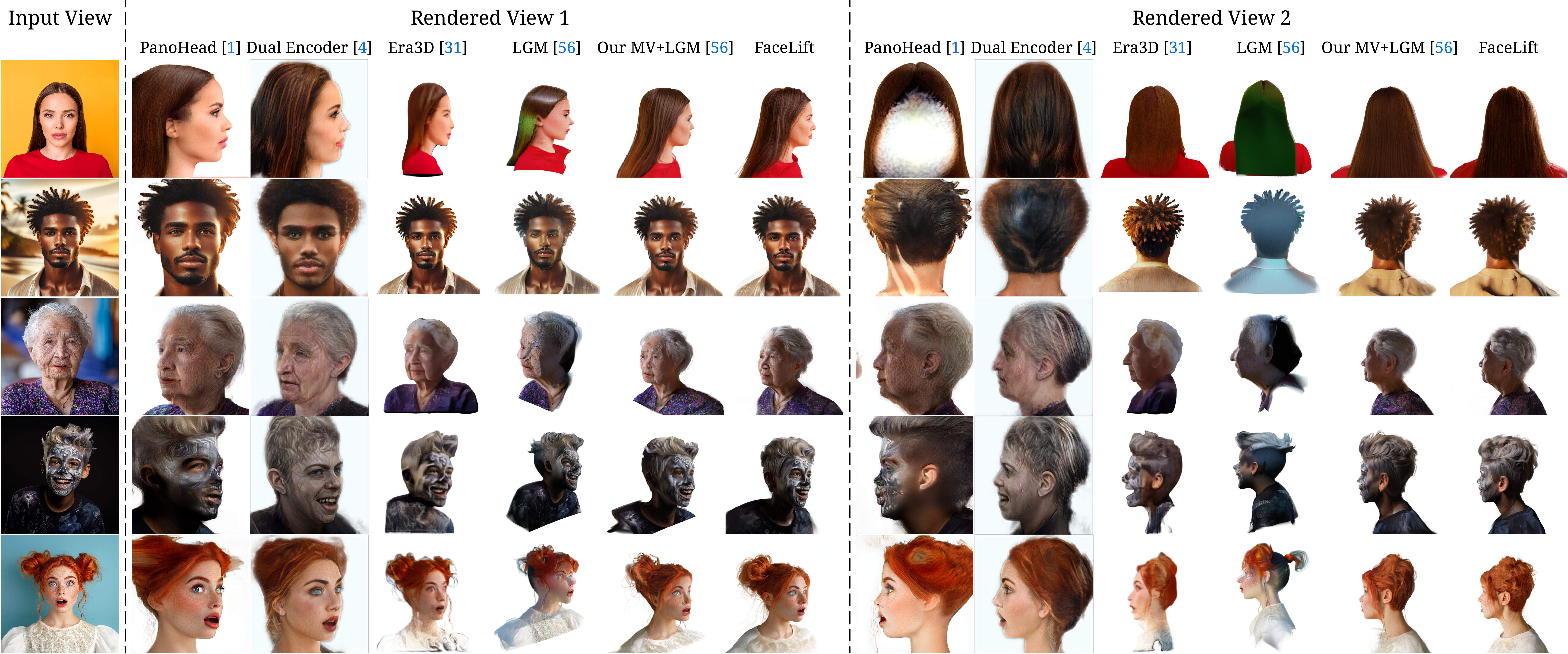}
    \vspace{-2mm}
    \caption{\textbf{Visual comparison on in-the-wild data.} {\ourmethod} demonstrates great generalization ability and robustness towards in-the-wild images, provides realistic unseen view rendering results. Era3D~\cite{era3d} and LGM~\cite{lgm} generate 3D head representation in inaccurate shape. PanoHead~\cite{panohead} often creates severe artifacts on the back of the head and can not handle challenging hairstyles well. Dual Encoder~\cite{dualencoder} shows improved performance on reconstructing the back of the head but exhibits more pronounced identity loss.}
    \label{fig:wild_compare}
    \vspace{-2mm}
\end{figure*}

\begin{figure*}[!ht]
    \centering
    \vspace{-2mm}
    \includegraphics[width=0.95\textwidth]{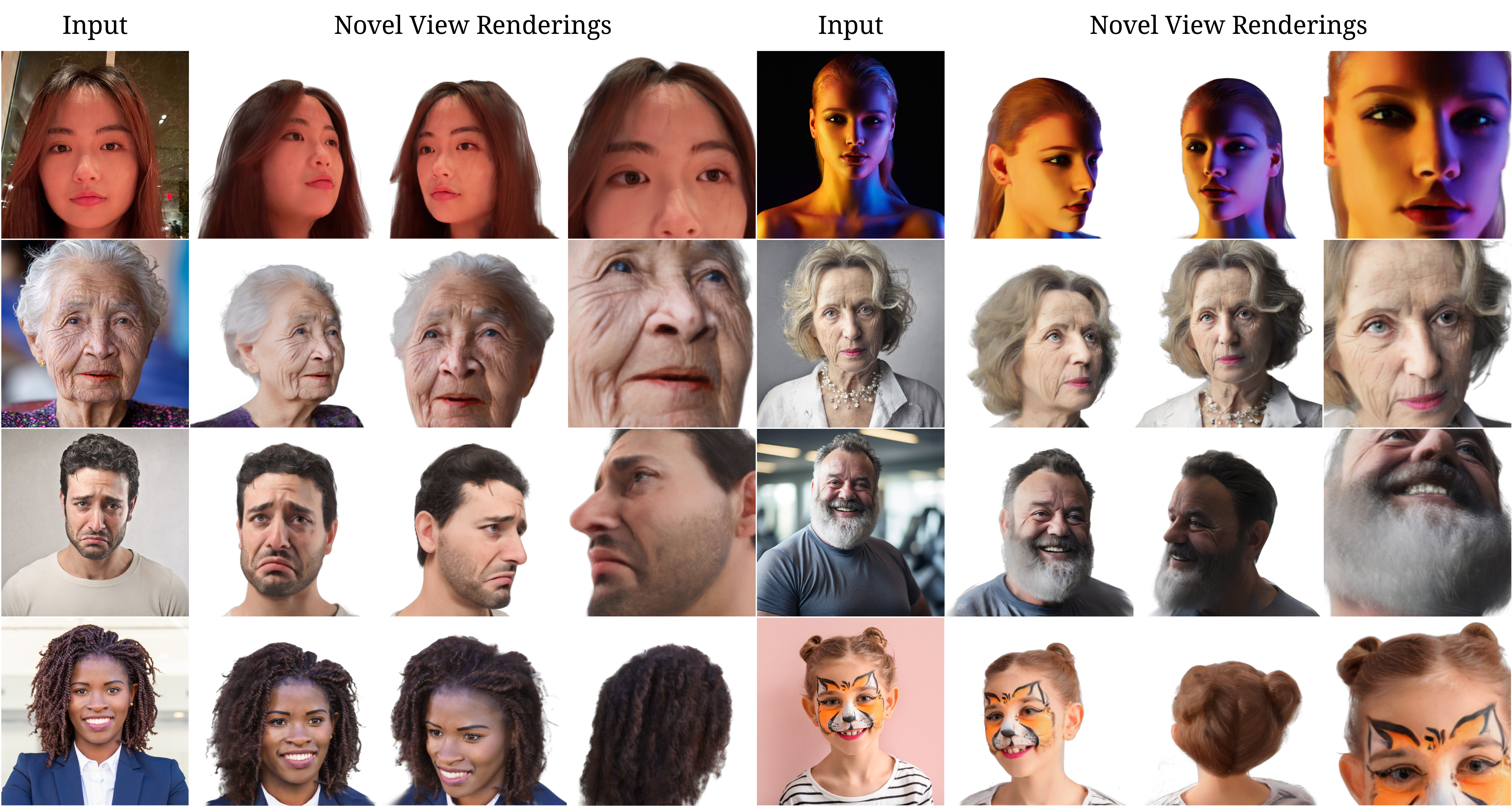}
    \vspace{-2mm}
    \caption{\textbf{Results of {\ourmethod} on in-the-wild images.} {\ourmethod} accurately reconstructs 3D head models under challenging lighting conditions, achieving high fidelity (row 1). It captures fine facial details such as wrinkles (row 2), mustaches (row 3), and individual hairs (row 2 and row 4). Additionally, it remains robust to complex facial expressions (row 3) and various skin tones (row 4). Furthermore, it can realistically reconstruct facial paint (row 4). More results are provided in the supplementary materials.}
    \label{fig:wild}
    \vspace{-4mm}
\end{figure*}

\vspace{-1mm}
\subsection{Experiments with In-the-wild Images}
\vspace{-1mm}
We collect in-the-wild human face images and present qualitative results in comparison with other baselines in Fig.~\ref{fig:wild_compare}.
%
Baseline methods often produce undesirable artifacts. For instance, PanoHead~\cite{panohead} frequently fails to render the back of the head and sometimes generates extra eyes at the rear. It also struggles to synthesize hair, shadows, wrinkles, and facial paint accurately, and its outputs lack multi-view consistency (\eg, the girl continues to face the camera in novel view 1 despite a changed pose).
Dual Encoder~\cite{dualencoder} improves back-of-head rendering but suffers from severe identity loss (row 2) and fails to accurately reconstruct face paint (row 4).
Era3D~\cite{era3d} often produces an inaccurate head shape, particularly from side views, and offers fewer geometric details compared to {\ourmethod}.
LGM~\cite{lgm} generates Gaussians with inaccurate color and opacity and lacks proper facial geometry, resulting in distorted features.
Baseline \textit{Our MV + LGM} shows that our multi-view diffusion model enhances LGM by providing improved facial geometry and texture details. However, the LGM reconstructor still produces Gaussians with inaccurate shapes and opacities, further underscoring the advantages of our transformer-based reconstructor.

We present more {\ourmethod}'s novel view rendering results in Fig.~\ref{fig:wild} to demonstrate {\ourmethod}'s ability to produce high-fidelity, realistic 3D head reconstructions with intricate details across a variety of challenging scenarios.
{\ourmethod} effectively handles faces under various lighting conditions. It can especially render realistic novel view images given a photo captured with an iPhone under dark lighting conditions (row 1 column 1), emphasizing its robustness and potential for real-world application.
It reconstructs facial details with high fidelity, especially the wrinkles and folds on the face caused by extreme expression.
{\ourmethod} also excels at reconstructing challenging textures, such as mustaches and hair.
Furthermore, it faithfully reconstructs facial paint, despite such data not being included in our synthetic face dataset, showcasing its strong generalization ability.




\vspace{-1mm}
\section{Ablation Study}
\label{sec:ablation_study}
\vspace{-1mm}
\begin{figure}[tb]
    \centering
    \includegraphics[width=0.47\textwidth]{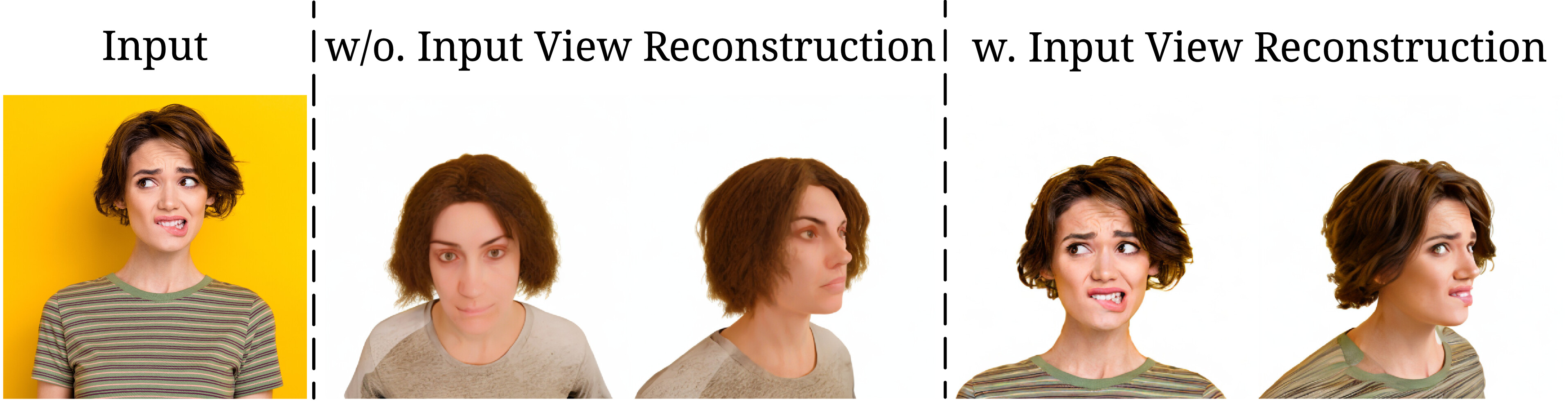}
    \vspace{-2mm}
    \caption{\textbf{Importance of input view reconstruction.} The diffusion model that is not trained to perform the input view reconstruction, \ie, \textit{w/o. Input View Reconstruction}, overfits to synthetic training distribution, suffers from severe identity loss during inference. Trained with input view reconstruction, our method preserves the input identity and expression faithfully.}
    \label{fig:input_view_reconstruction}
    \vspace{-4mm}
\end{figure}
\subsection{Input View Reconstruction}
\vspace{-1mm}
\label{subsec:input_view_reconstruction}
We conduct an ablation study to demonstrate the importance of reconstructing the input view during training.
For comparison, we train a multi-view diffusion model that generates six novel views. In this baseline, the first generated view’s elevation is adjusted from \(0^{\circ}\) to \(20^{\circ}\), while the remaining views adopt the same camera poses as in our default setting. We refer to this variant as \textit{w/o. Input View Reconstruction}.
Fig.~\ref{fig:input_view_reconstruction} presents the view generation results of the two diffusion models when applied to real-world images. Without the input view reconstruction task, the model trained on the synthetic dataset generates views within a limited distribution, leading to noticeable identity loss. Moreover, it loses its ability to preserve facial expressions and face paint. In contrast, incorporating the input view reconstruction task during training enables our diffusion model to faithfully regenerate the input view, significantly improving its generalization ability. Quantitative comparison is provided in the supplementary material.


\vspace{-1mm}
\subsection{Number of Views}
\begin{figure}[tb]
    \centering
    \includegraphics[width=.47\textwidth]{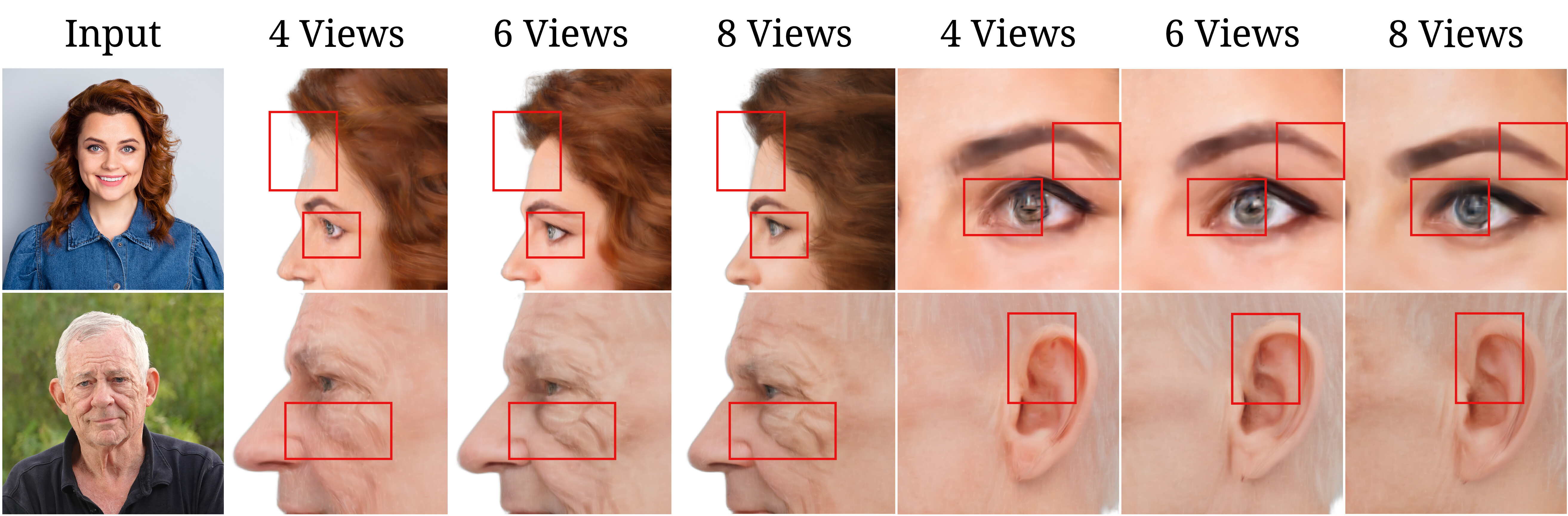}
    \vspace{-2mm}
    \caption{\textbf{Number of input views of Gaussian reconstructor.} Using six views strike a good balance between reconstruction quality and computational efficiency.}
    \label{fig:ablation_view}
    \vspace{-4mm}
\end{figure}

\vspace{-1mm}
\label{subsec:num_of_views}
We evaluate three configurations: four views (front, left, right, and back), six views (adding front-left and front-right), and eight views (further including front-top and front-bottom). Fig.~\ref{fig:ablation_view} compares the baselines using different numbers of input views. With only four views, the reconstructor fails to capture a complete forehead; however, with six views, it reconstructs the eyes and eyebrows more smoothly and renders challenging textures—such as facial wrinkles and ear folds—more realistically. Eight views do not offer significant visual improvements, and incur a higher computational cost in both stages. Thus, we find that six views achieve a good balance between reconstruction quality and computational efficiency.

\begin{figure}[tb]
    \centering
    \includegraphics[width=.47\textwidth]{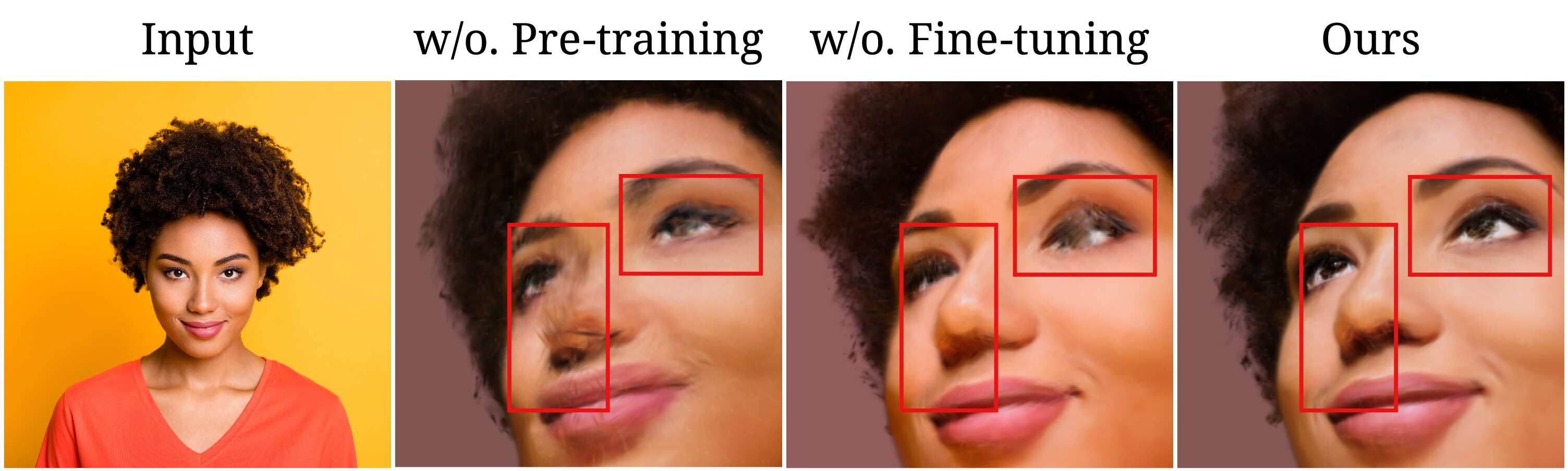}
    \vspace{-2mm}
    \caption{\textbf{Two-stage training of reconstructor.} Without pre-training on general objects, the reconstructor fails to produce clear textures in the reconstruction results. Meanwhile, without fine-tuning on synthetic human head data, the model lacks a refined understanding of facial structures, including the eyes and nose.}
    \label{fig:lrm_training}
    \vspace{-4mm}
\end{figure}
\vspace{-1mm}
\subsection{Two-stage Reconstructor Training}
\vspace{-1mm}
\label{subsec:two_stage_reconstructor_training}
As illustrated in Sec.~\ref{subsec:gs-lrm}, our Gaussian reconstructor follows a two-stage training pipeline. Fig.~\ref{fig:lrm_training} shows that pre-training on general objects helps the model learn a diverse prior of geometric structures, resulting in clearer textures on delicate facial regions. Meanwhile, fine-tuning on synthetic human head data enables the reconstructor to gain a more refined understanding of facial structure, thereby enhancing the accuracy of features such as the eyes, nose, and hair.

\vspace{-1mm}
\section{Conclusions}
\vspace{-1mm}
\label{sec:conclusions}
We propose {\ourmethod}, a feed-forward approach that lifts a single facial image to a detailed 3D reconstruction with preserved identity features. Our method uses multi-view diffusion to generate unobservable views and employs a transformer-based reconstructor to reconstruct 3D Gaussian splats, enabling high-quality novel view synthesis. To overcome the difficulty of capturing real-world multi-view human head images, we render high-quality synthetic data for training and show that, despite being trained solely on synthetic data, {\ourmethod} can reconstruct 3D heads from real-world captured images with high fidelity. Compared with baselines~\cite{panohead, dualencoder, gghead, era3d, lgm}, {\ourmethod} generates 3D head representation with finer geometry and texture details and exhibits better identity preservation ability. 

\clearpage
\section{Acknowledgement}
This was supported in part by the Institute of Information \& Communications Technology Planning \& Evaluation (IITP) grant funded by the Korean Government (MSIT) (No. RS-2024-00457882, National AI Research Lab Project).

We appreciate the insightful discussions with Kai Zhang, Hao Tan, Zexiang Xu, Sai Bi, Sumit Chaturvedi, Hanwen Jiang, Yu-Ju Tsai, Kuan-Chih Huang, Chengxu Liu, and Dingyi Dai. We thank Nathan Carr and Kalyan Sunkavalli for their support.

{
    \small
    \bibliographystyle{ieeenat_fullname}
    \bibliography{main}
}
\clearpage
\setcounter{section}{0}
\maketitlesupplementary

\section{Overview}
This supplementary material presents additional results to complement the main manuscript.
We first provide a supplementary video showcasing additional visual results.
We then provide further experiments in Sec.~\ref{sec:supp_exp}, including a comparison with DimensionX~\cite{dx}, additional visual results of {\ourmethod} on in-the-wild images, additional ablation study results and an autoregressive generation pipeline to apply {\ourmethod} on videos to achieve 4D rendering.
We deliver more details on our method in Sec.~\ref{sec:supp_method_details} and illustrate experimental details in Sec.~\ref{sec:supp_exp_details}.
Finally, we discuss the limitations of {\ourmethod} in Sec.~\ref{sec:supp_limitations}.

\section{Supplementary Video}
Please refer to our supplementary video for a more comprehensive visualization of the results. The video includes additional examples of single-image-to-3D head reconstruction, demonstrations in the interactive viewer, and results showcasing video-based input for 4D novel view synthesis.

\section{Additional Experiments}
\label{sec:supp_exp}
\subsection{Comparison with DimensionX}
\begin{figure}[tb]
    \centering
    \includegraphics[width=.47\textwidth]{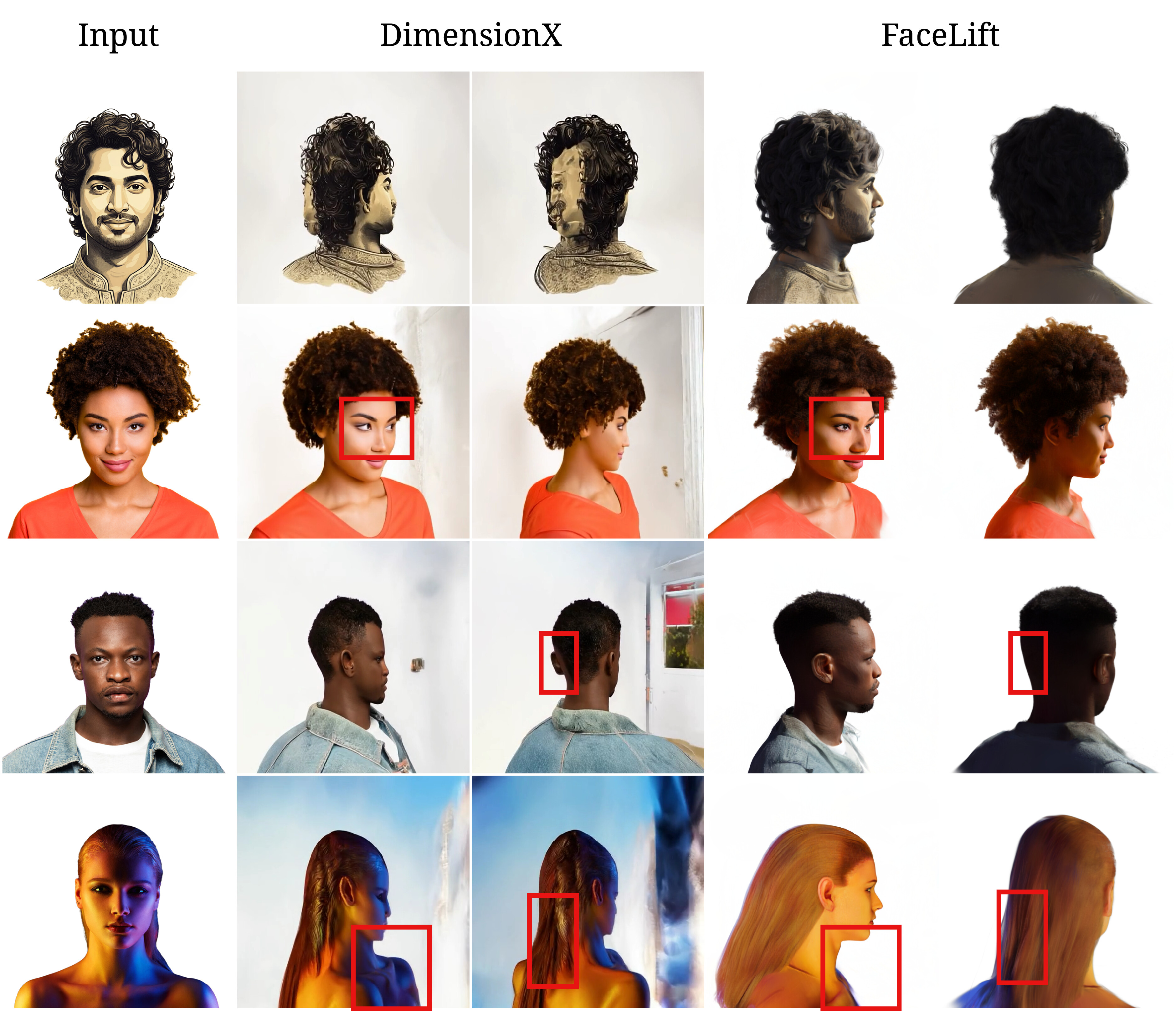}
    \caption{\textbf{Visual comparison with DimensionX~\cite{dx}.} DimensionX frequently produces inaccuracies in the back of the head and the shoulder shapes. Other common issues include misaligned ears and eyes gazing in incorrect directions. Additionally, controlling camera poses is challenging. In contrast, {\ourmethod} delivers results that are significantly more consistent across multiple views while enabling the generation of more visually appealing hair.}
    \label{fig:supp_dx}
\end{figure}
We provide additional comparison results on single image to 3D tasks with a state-of-the-art video diffusion model, DimensionX~\cite{dx}. DimensionX is a framework designed to generate photorealistic 3D and 4D scenes from a single image with video diffusion. The results are shown in Fig.~\ref{fig:supp_dx}. As a video diffusion model, DimensionX struggles to produce multi-view consistent results and lacks a clear spatial understanding of head shapes. As a result, it often generates eyes gazing in the wrong direction and ears positioned incorrectly, along with inaccurate shoulder shapes. In contrast, {\ourmethod} generates highly realistic 3D human heads while also producing more visually striking hair.

\subsection{Comparison with Mesh-based Methods}
\begin{figure}[tb]
    \centering
    \includegraphics[width=.47\textwidth]{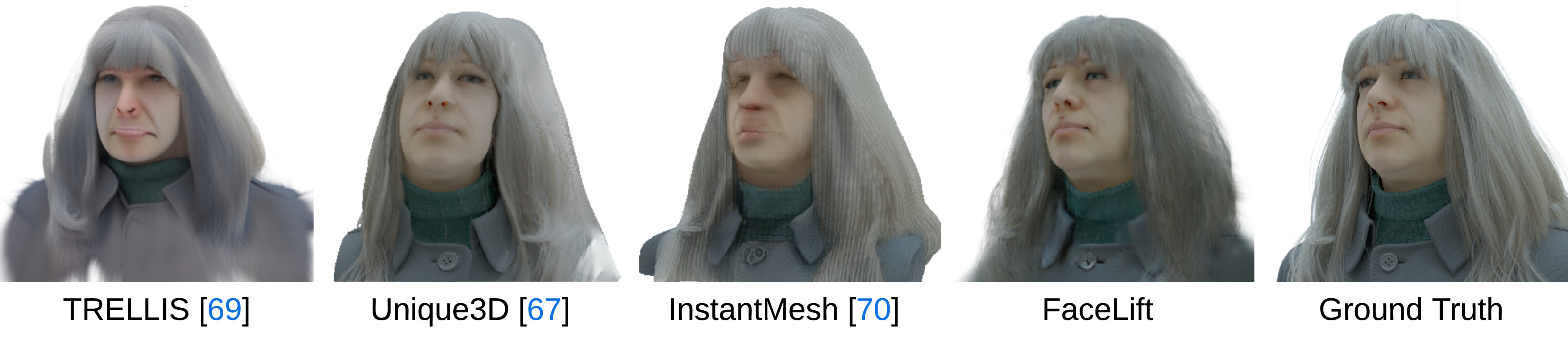}
    \caption{\textbf{Visual results on Cafca compared with mesh-based reconstruction methods.} Compared to mesh-based reconstruction methods, our use of pixel-aligned 3D Gaussians offers clear advantages: the semi-transparent kernels naturally capture complex visual phenomena such as hair strands and fine wrinkles.}
    \label{fig:rebuttal_fig}
\end{figure}

We provide comparison results with mesh-based reconstruction methods InstantMesh~\cite{instantmesh}, Unique3D~\cite{unique3d}, and TRELLIS~\cite{trellis} on the Cafca dataset~\cite{cafca}. Quantitative results are shown in Tab.~\ref{tab:rebuttal}, and quantitative comparisons are shown in Fig.~\ref{fig:rebuttal_fig}. Results show that mesh-based reconstruction methods fail to provide realistic hair texture and detailed skin wrinkles. Meanwhile, thanks to the input view reconstruction strategy, {\ourmethod} achieves superior identity preservation.

\begin{table}[tb]
    \centering
    \scriptsize
    \begin{tabular}{@{\hskip 2.0mm}l@{\hskip 2.0mm}c@{\hskip 2.0mm}c@{\hskip 2.0mm}c@{\hskip 2.0mm}c@{\hskip 2.0mm}c@{\hskip 2.0mm}}
        \toprule
        Method & PSNR $\uparrow$ & SSIM $\uparrow$ & LPIPS $\downarrow$ & DreamSim $\downarrow$ & ArcFace $\downarrow$ \\
        \midrule
        TRELLIS~\cite{trellis} & 12.74 & 0.7412 & 0.3746 & 0.2170 & 0.4001 \\
        Unique3D~\cite{unique3d} & 14.27 & 0.7643 & 0.3188 & 0.1277 & 0.2088 \\
        InstantMesh~\cite{instantmesh} & 16.44 & 0.7815 & 0.2792 & 0.1504 & 0.2741 \\
        {\ourmethod} & \textbf{16.61} & \textbf{0.7968} & \textbf{0.2694} & \textbf{0.1096} & \textbf{0.1573} \\
        \bottomrule
    \end{tabular}
    \caption{\textbf{Quantitative results on Cafca compared with mesh-based reconstruction methods.} {\ourmethod} achieves better quantitative results with more suitable representations and specialized training data.}
    \label{tab:rebuttal}
\end{table}

\subsection{Additional Results on In-the-wild Images}
\begin{figure*}[!ht]
    \centering
    \includegraphics[width=0.94\linewidth]{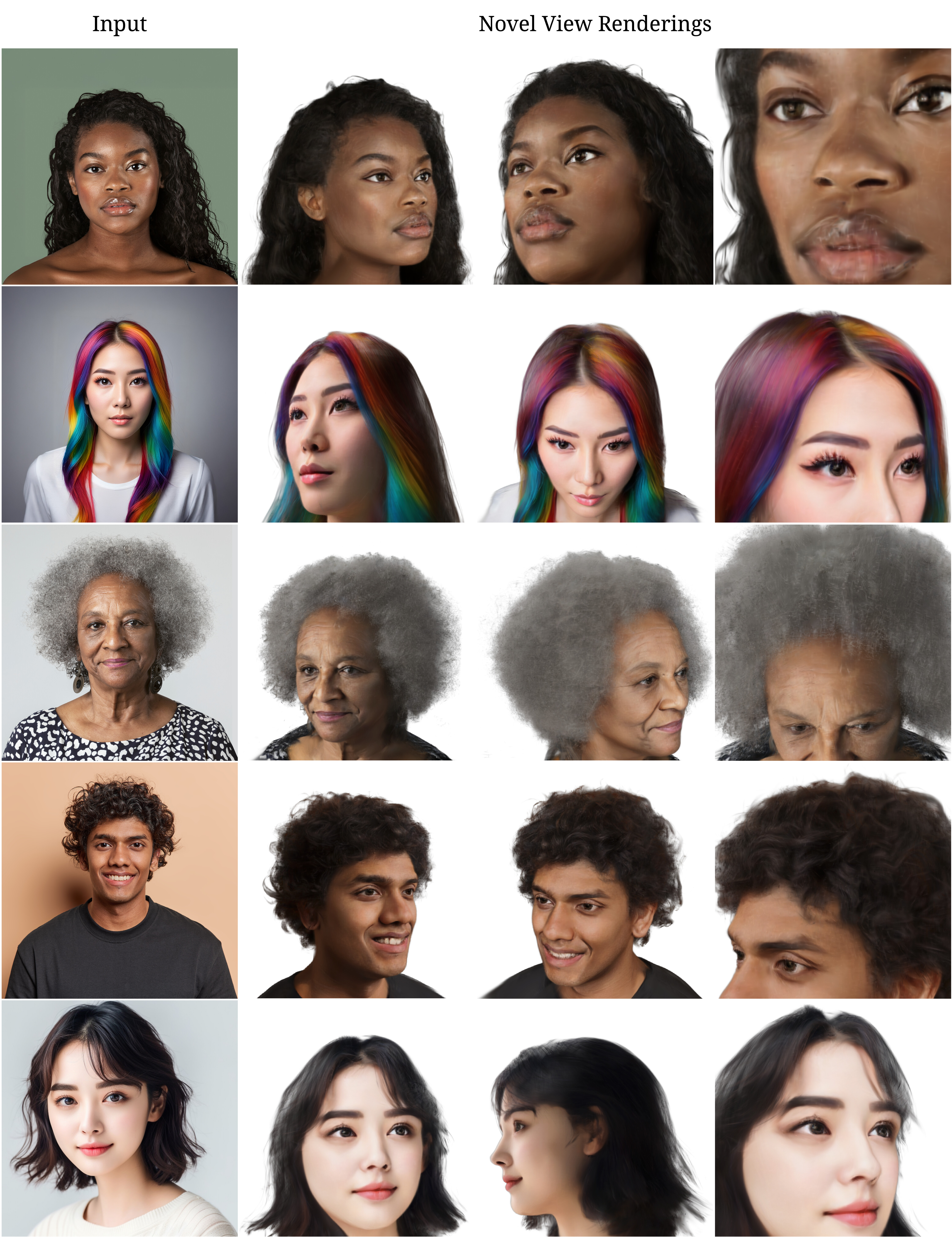}
    \caption{\textbf{Results of {\ourmethod} on in-the-wild images.} {\ourmethod} excels at reconstructing intricate and diverse facial hair, encompassing a wide array of hairstyles and hair colors. It also accurately captures a broad range of skin tones.}
    \label{fig:supp_wild_1}
\end{figure*}

\begin{figure*}[!ht]
    \centering
    \includegraphics[width=0.94\linewidth]{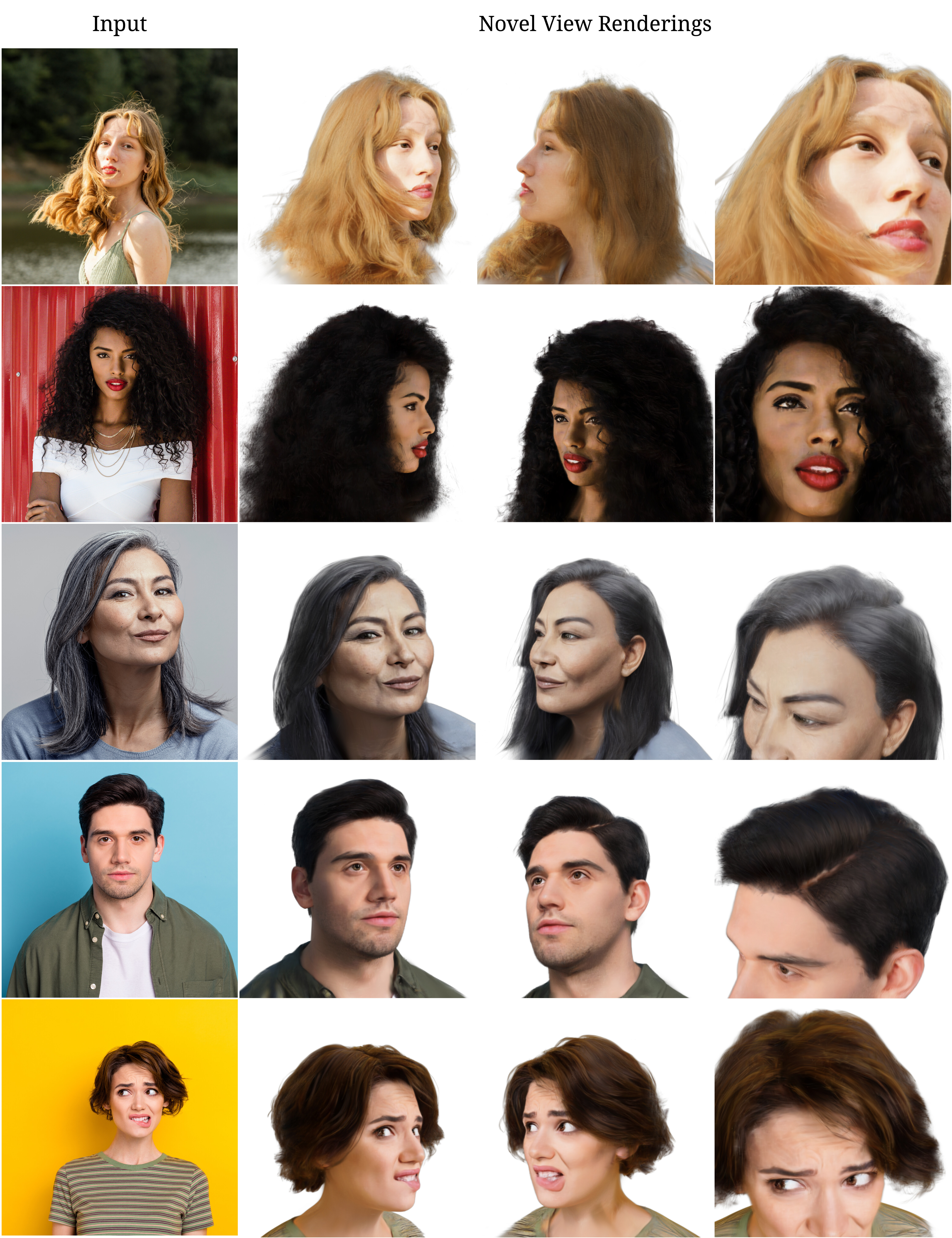}
    \caption{\textbf{Results of {\ourmethod} on in-the-wild images.} {\ourmethod} also demonstrates the ability to reconstruct faces exhibiting a wide range of pose variations. It can also accurately handle extreme expressions.}
    \label{fig:supp_wild_2}
\end{figure*}

\begin{figure*}[!ht]
    \centering
    \includegraphics[width=0.94\linewidth]{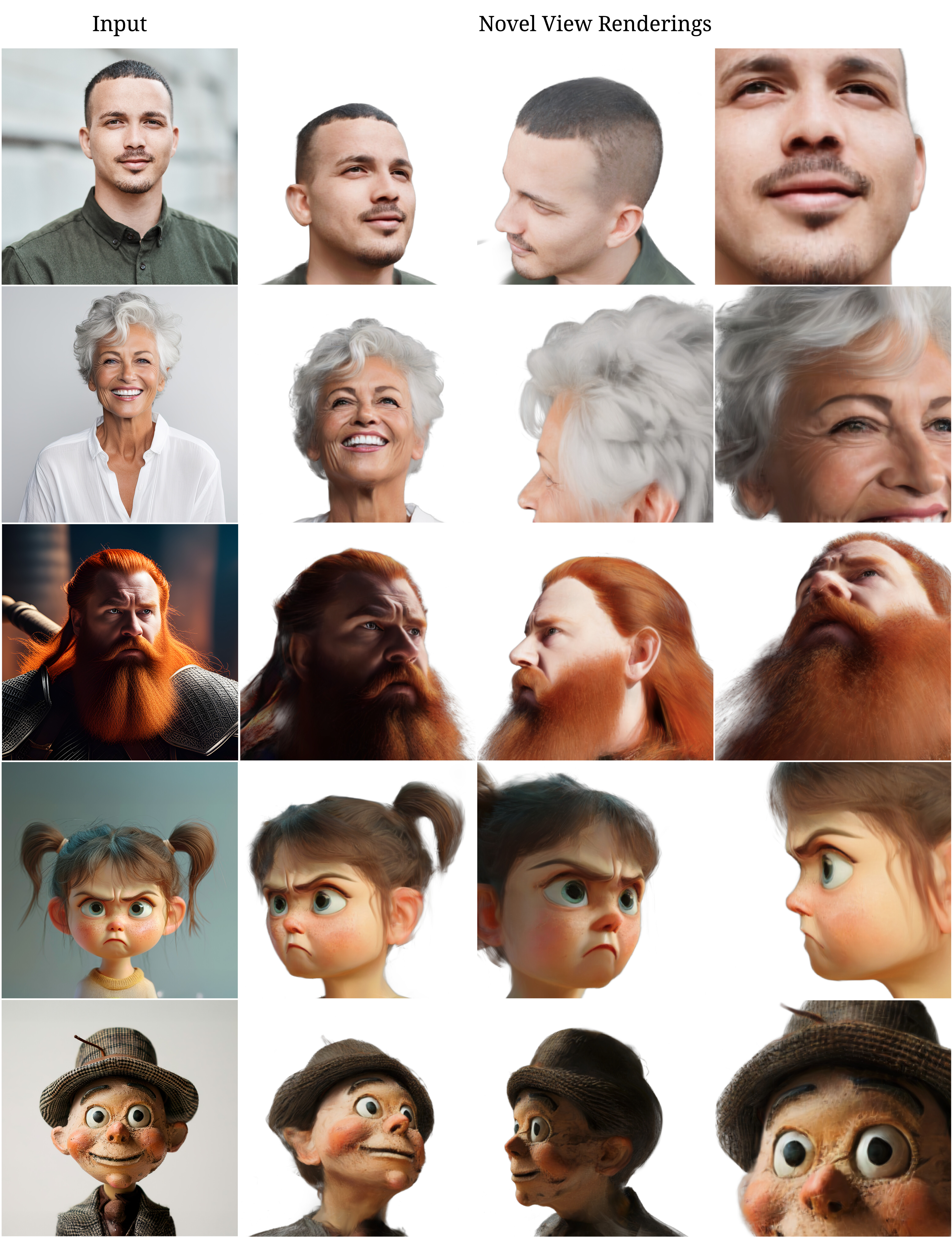}
    \caption{\textbf{Results of {\ourmethod} on in-the-wild images.} {\ourmethod} realistically reconstructs detailed facial textures. Additionally, {\ourmethod} is well-suited for reconstructing cartoon characters.}
    \label{fig:supp_wild_3}
\end{figure*}

We present additional results on in-the-wild images in Fig.~\ref{fig:supp_wild_1}, Fig.~\ref{fig:supp_wild_2} and Fig.~\ref{fig:supp_wild_3}. {\ourmethod} demonstrates the ability to effectively handle diverse hairstyles and beards. Notably, it excels at hallucinating unobservable hairline splits and synthesizing the transparent properties of hair using Gaussians with low opacity. Our method reconstructs photo-realistic 3D heads under various lighting conditions and can be further extended to the reconstruction of cartoon characters.

\subsection{Additional Ablation Study}
\begin{figure}[tb]
    \centering
    \includegraphics[width=.47\textwidth]{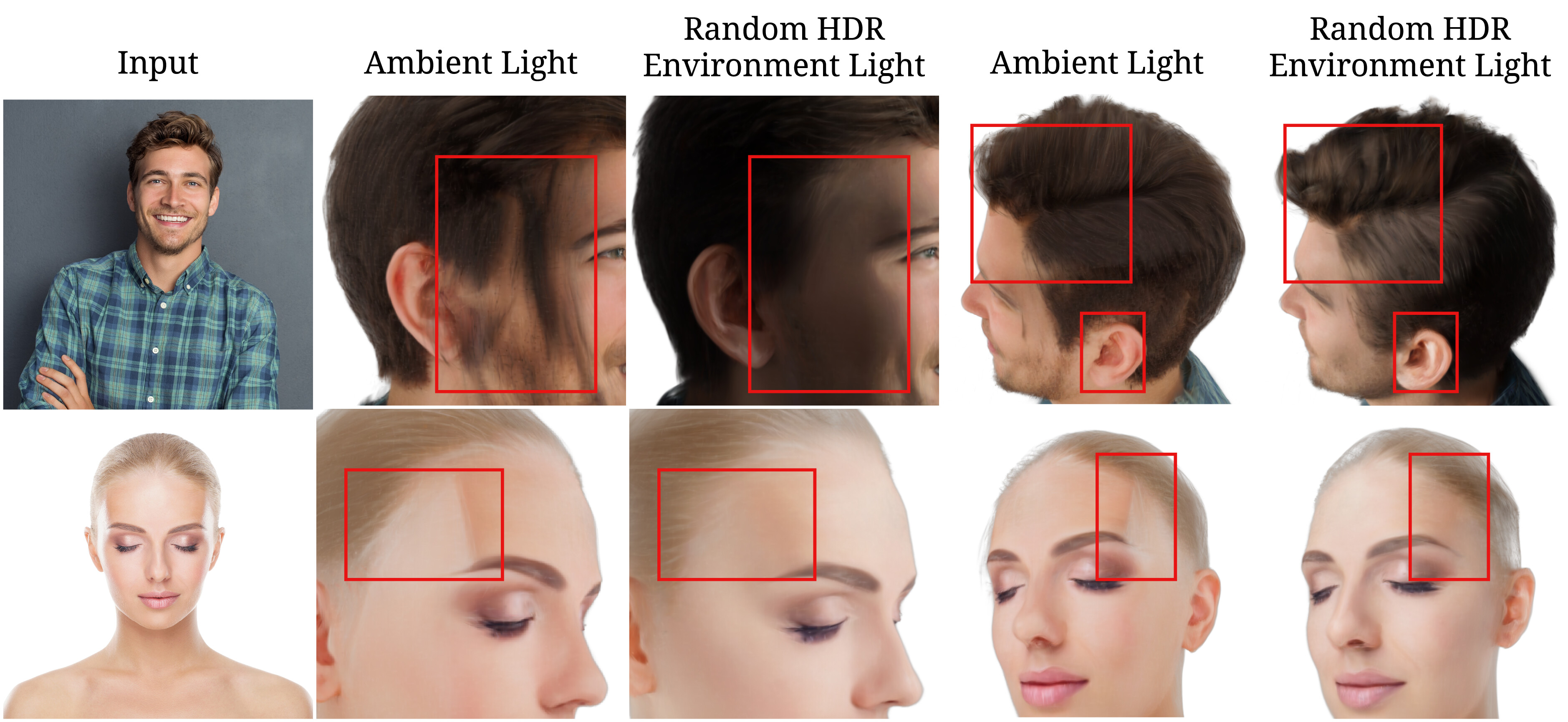}
    \caption{\textbf{Ablation study on synthetic data lighting condition.} Models trained only with ambient light struggle to handle shadows and strong lighting.}
    \label{fig:ablation_light}
\end{figure}
\begin{figure*}[tb]
    \centering
    \includegraphics[width=0.95\linewidth]{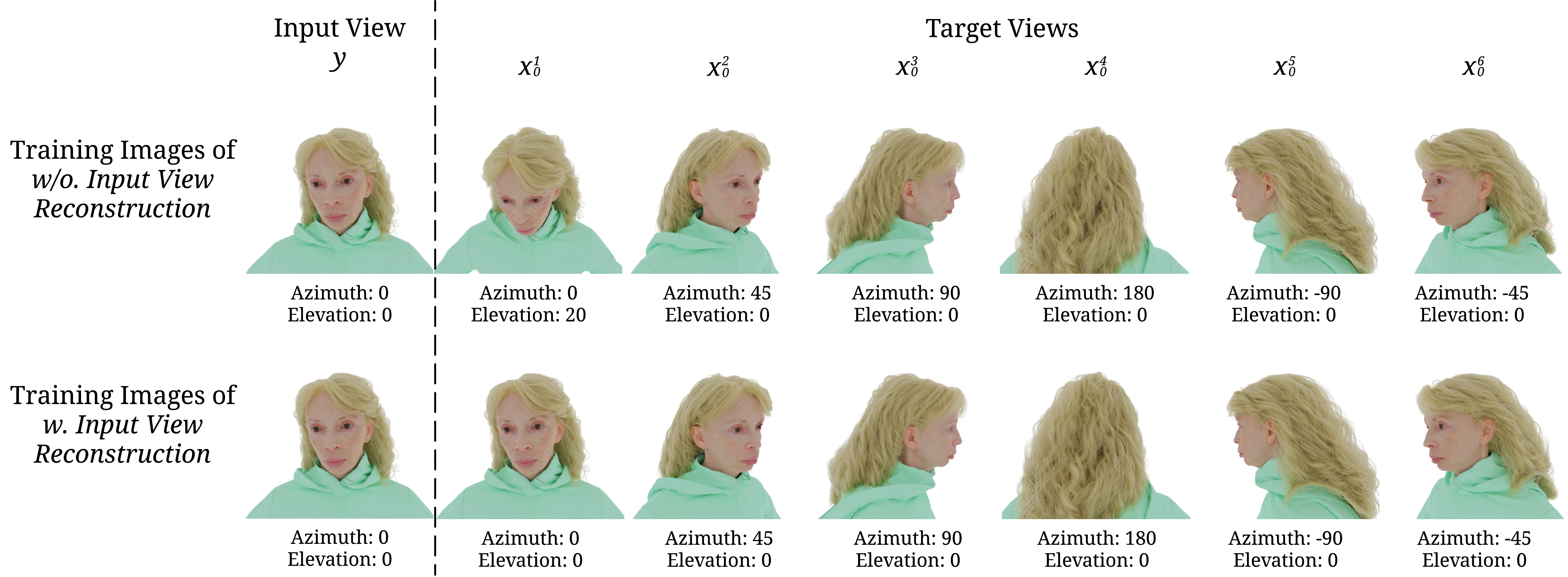}
    \caption{\textbf{Training images used in the study of input view reconstruction.} We show example images for training baselines \textit{w/o. Input View Reconstruction} and \textit{w. Input View Reconstruction}. The difference lies in the elevation of the first target image.}
    \label{fig:supp_training_images}
\end{figure*}
\begin{figure*}[tb]
    \centering
    \includegraphics[width=0.95\linewidth]{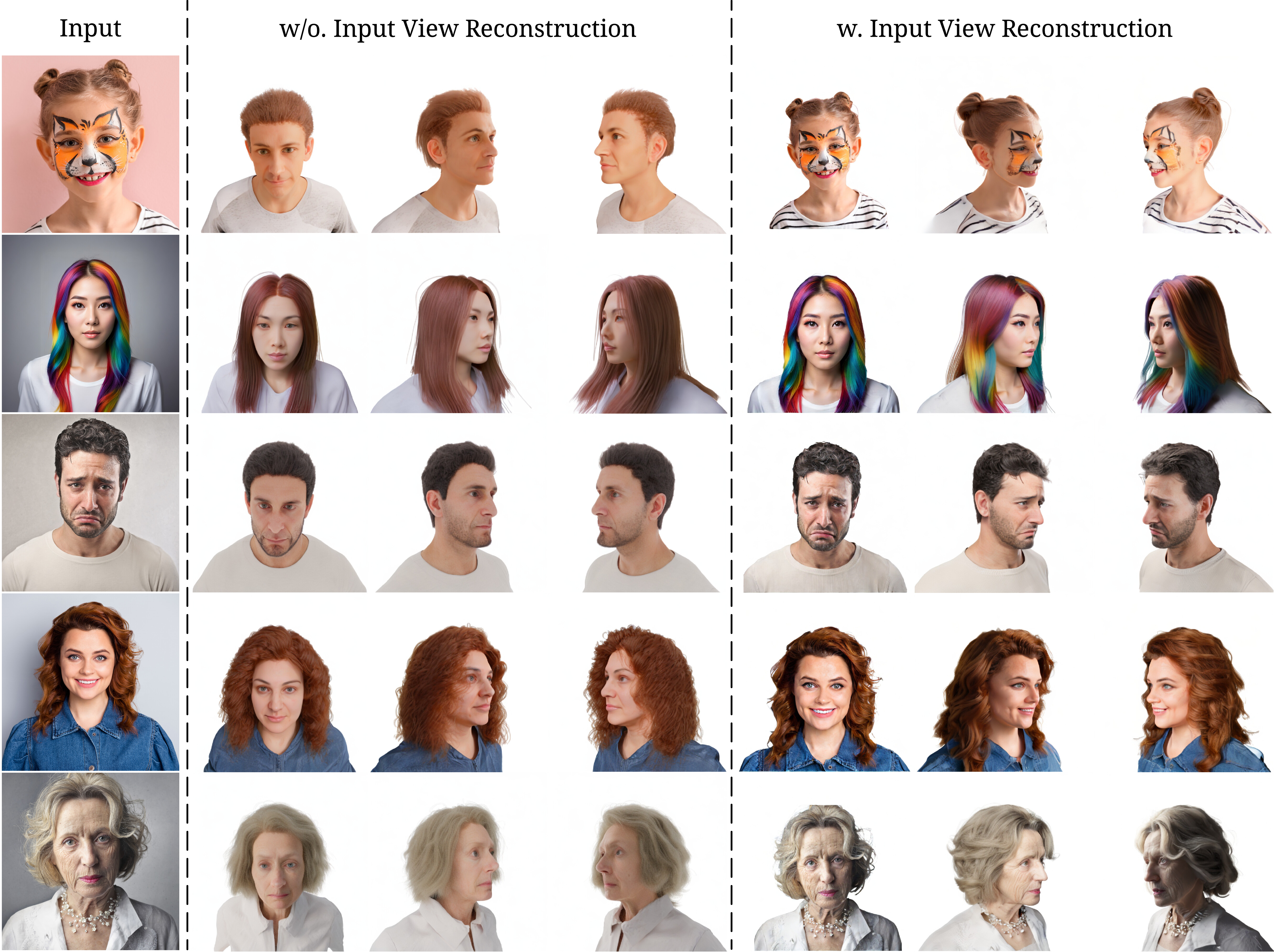}
    \caption{\textbf{Importance of input view reconstruction.} The diffusion model without input view reconstruction training suffers from identity loss. Additionally, it fails to generate accurate face paint (row 1), diverse hair colors (row 2), varied expressions (row 3 and 4), and accessories (row 5).}
    \label{fig:supp_input_view_reconstruction}
\end{figure*}
\noindent\textbf{Importance of Data with Diverse Lighting.}
We use synthetic data to train our models, which offers the advantage of controlling lighting conditions and rendering head images under various lighting scenarios. In contrast, real-world human data is typically captured in a studio with lighting similar to ambient light, as shown in the input of Fig.~\ref{tab:ava_256}. To highlight the importance of training models with diverse lighting conditions, we train {\ourmethod} with (1) Data rendered with only ambient light, and (2) Data rendered in random HDR environment light. We present the visual result comparison in Fig.~\ref{fig:ablation_light}. The model trained exclusively on ambient light data struggles to understand shadows, often generating hair-like textures on the face. Furthermore, when exposed to strong light, it produces white regions on the face. In contrast, the model trained with random HDR environment light generates smooth transitions between regions with different lighting conditions.

\vspace{1mm}
\noindent\textbf{More Results on Input View Reconstruction.}
\begin{table}[tb]
    \centering
    \scriptsize
    \begin{tabular}{@{\hskip 0.7mm}l@{\hskip 0.7mm}c@{\hskip 0.7mm}c@{\hskip 0.7mm}c@{\hskip 0.7mm}c@{\hskip 0.7mm}c@{\hskip 0.7mm}}
        \toprule
        Baseline & PSNR $\uparrow$ & SSIM $\uparrow$ & LPIPS $\downarrow$ & DreamSim $\downarrow$ & ArcFace $\downarrow$ \\
        \midrule
        w/o Input View Reconstruction & 16.02 & 0.7884 & 0.2893 & 0.1438 & 0.2367 \\
        w/o Multi-view Attention & 16.29 & 0.7885 & 0.2861 & 0.1552 & 0.2126 \\
        Full Model & \textbf{16.61} & \textbf{0.7968} & \textbf{0.2694} & \textbf{0.1096} & \textbf{0.1573} \\
        \bottomrule
    \end{tabular}
    \caption{\textbf{Quantitative results of ablation studies.} {\ourmethod} achieves better quantitative results with more suitable representations and specialized training data.}
    \label{tab:rebuttal_2}
\end{table}
\vspace{1mm}
\noindent\textbf{}
We show training samples for two baselines \textit{w/o. Input View Reconstruction} and \textit{w. Input View Reconstruction} in Fig.~\ref{fig:supp_training_images}. As the target views are different, baseline \textit{w/o. Input View Reconstruction} is trained to generate six images with novel camera poses, while baseline \textit{w. Input View Reconstruction} reconstruct the input image and generate five images with novel poses. Inference results on real world images are displayed in Fig.~\ref{fig:supp_input_view_reconstruction} to illustrate the importance of reconstructing the input image during multi-view diffusion training. The results demonstrate that input view regeneration prevents the model from being confined to the training data distribution, thereby enhancing its ability to preserve identity. Quantitative results of baseline \textit{w/o. Input View Reconstruction} is shown in Tab.~\ref{tab:rebuttal_2}.

\begin{figure}[tb]
    \centering
    \includegraphics[width=1.0\linewidth]{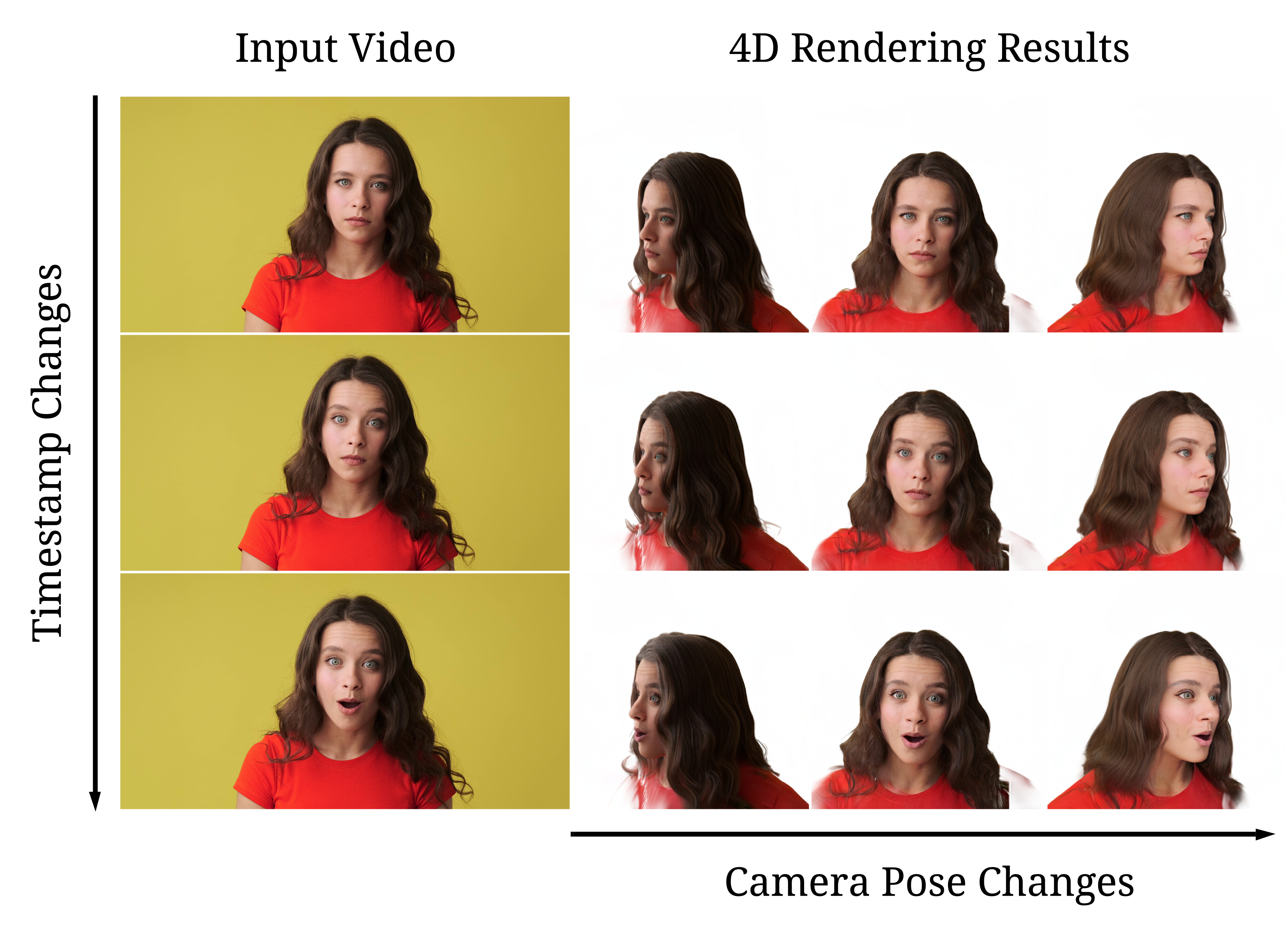}
    \caption{\textbf{Results of directly applying {\ourmethod} to video input.} By processing each video frame independently, {\ourmethod} generates a sequence of Gaussians that preserves consistent visual identity and accurately captures facial expressions. However, this baseline does not consider temporal consistency.}
    \label{fig:supp_direct_4d_renderings}
    \vspace{-3mm}
\end{figure}
\begin{figure*}[ht]
    \centering
    \includegraphics[width=0.95\textwidth]{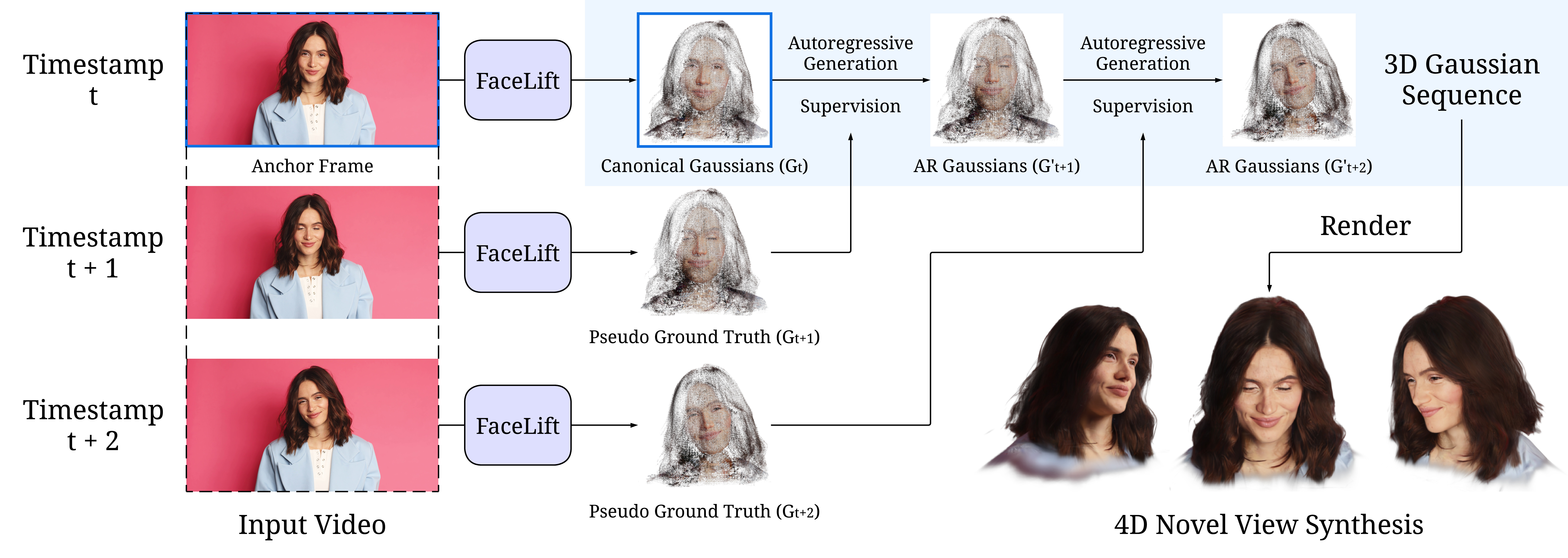}
    \caption{\textbf{Autoregressive Generation for 4D Rendering.} "AR Gaussians" denotes autoregressively generated Gaussians. With {\ourmethod}, each video frame is independently converted into a 3D Gaussian representation. An anchor frame at timestamp $t$ (highlighted by the blue box) produces Canonical Gaussians $G_t$, which are then deformed into the representations for subsequent frames, $G'_{t+1}$, $G'_{t+2}$, \dots, \etc. This deformation is supervised by the rendered output Gaussians $G_{t+1}$, $G_{t+2}$, \dots, \etc, produced by {\ourmethod}. Iteratively applying this process yields a temporally consistent Gaussian sequence that supports rendering from any viewpoint.}
    \label{fig:supp_auto_gen}
\end{figure*}
\begin{figure}[t]
    \centering
    \includegraphics[width=1.0\linewidth]{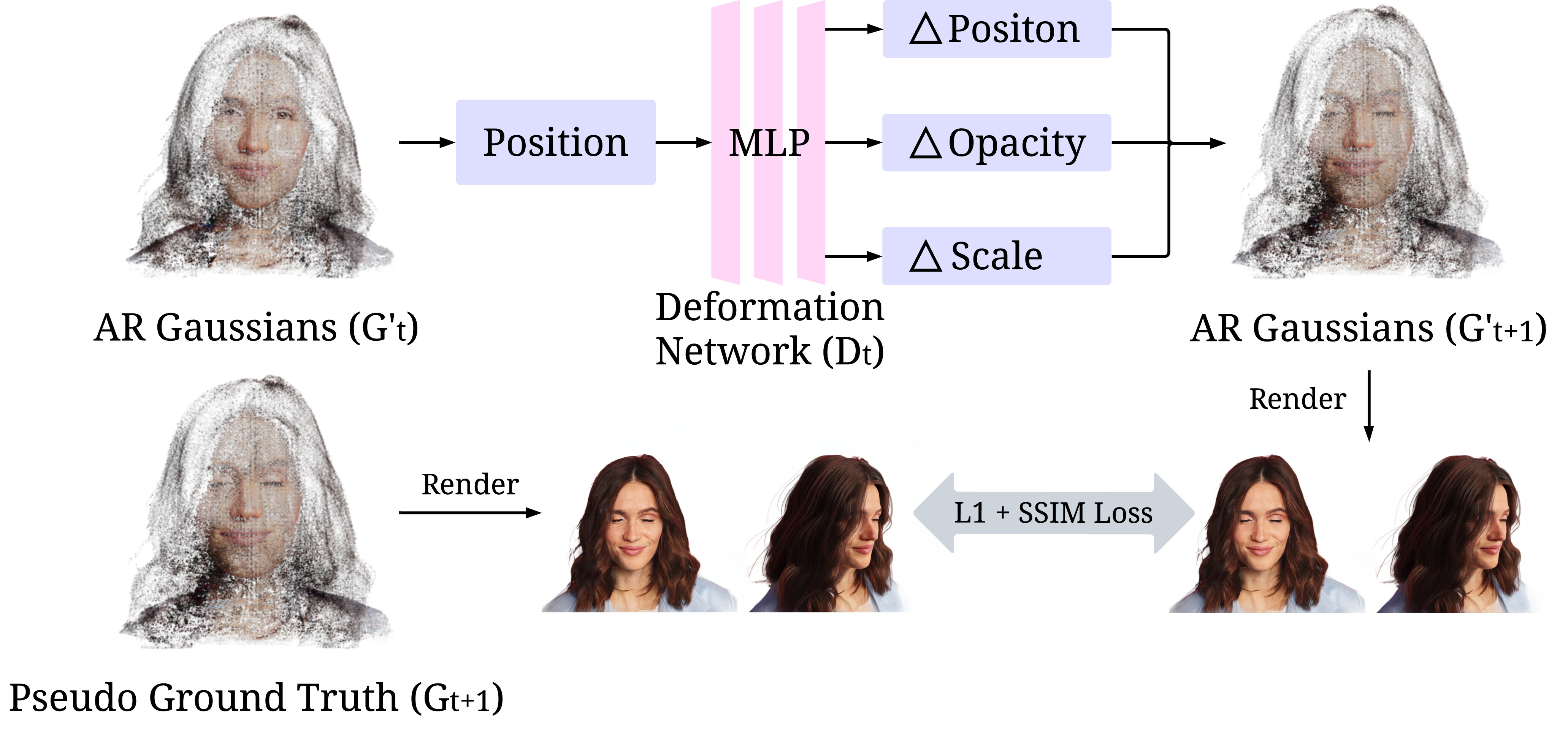}
    \caption{\textbf{Deformation Network.} The deformation network $D_t$ is an eight-layer MLP that predicts geometric deformations, including positional shifts, opacity adjustments, and scale changes. Combined with the Gaussian representations from the previous frame $G_t'$, it forms the Gaussian representation for the next frame $G_{t+1}'$.}
    \label{fig:supp_deformation}
\end{figure}
\begin{figure*}[t]
    \centering
    \includegraphics[width=0.95\linewidth]{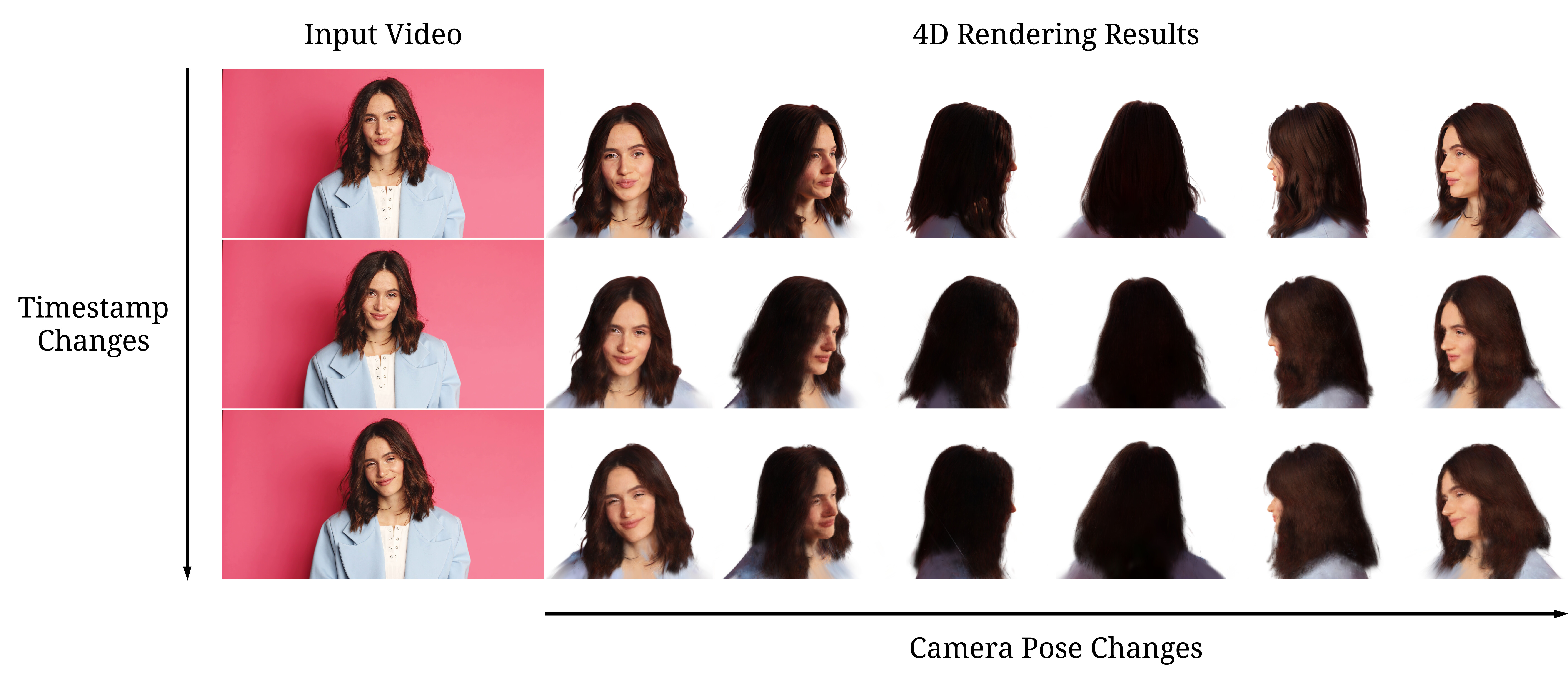}
    \caption{\textbf{Results of applying {\ourmethod} on video.} Our proposed autoregressive generation pipeline enables {\ourmethod} to be applied directly to video sequences, achieving 4D novel view synthesis -- rendering at any given timestamp and camera pose. Video results are shown in the supplementary material.}
    \label{fig:supp_4d_renderings}
\end{figure*}
\subsection{Applying {\ourmethod} on Videos}
\label{subsec:4d}
{\ourmethod} can be directly applied to video frames and achieve high-quality facial reconstructions with consistent visual identity and accurate facial expression, as shown in Fig.~\ref{fig:supp_direct_4d_renderings}. However, since {\ourmethod} is not trained on video data, many full-head details are generated independently by the diffusion models, resulting in subtle flickering. In this supplemental document, we introduce a simple yet effective method that leverages {\ourmethod} and autoregressive training to achieve high-quality, temporally smooth 4D facial reconstructions.

Given an input video $\{F_0, F_1, \dots F_T\}$, we process each video frame $F_t$ sequentially to generate a set of 3D Gaussian sequences $\{G_0, G_1, \dots G_T\}$, where each $G_t$ represents the obtained Gaussian representation at timestamp $t$. As each $G_t$ is generated from frame $I_t$ without interaction with other frames, directly rendering from this Gaussian sequence creates artifacts resulting from time-inconsistency. Hence, we propose an autoregressive generation pipeline, as shown in Fig.~\ref{fig:supp_auto_gen}.

We first select an anchor frame at timestamp $t$ (marked with blue box), and treat its corresponding 3D Gaussian splats as the canonical Gaussians $G_t$ (marked with blue box). Then, for a following timestamp $t+1$, we train a deformation network $D_t$ to predict Gaussian splats $G_{t+1}'$ deformed from $G_t$ supervised by rendering results from $G_t$. The deformation network is an 8-layer MLP, which takes the $x, y, z$ position of each Gaussian in $G_t$ as input and predicts $\Delta x$, $\Delta y$, $\Delta z$, opacity change $\Delta\alpha$ and scale change $\Delta s$. These deformation parameters are combined with $G_t$ to generate $G_{t+1}'$, as shown in Fig.~\ref{fig:supp_deformation}.

To train the deformation network, we render six views with the same camera poses as the multi-view diffusion outputs from $G_{t+1}'$, and the renderings of the same camera poses from $G_{t+1}$ are used as pseudo ground truth supervision. Then we treat $G_{t+1}'$ as the initial Gaussians and train deformation network $D_{t+1}$ to generate $G_{t+2}'$. Iteratively, we will get a Gaussian sequence $\{G_0', G_1', \dots, G_T'\}$. Given any timestamp, we can select the corresponding 3D Gaussians from this Gaussian sequence and render from any given pose. The results of this method are shown in Fig.~\ref{fig:supp_4d_renderings}, which demonstrate improved temporal consistency while preserving identity and achieving accurate expression modeling. Please refer to the supplementary video for additional video rendering results.

\section{Method Details}
\label{sec:supp_method_details}
\subsection{Details on View Generation}
Given a single near frontal view face image with azimuth $\alpha$, the multi-view diffusion model will generate six views with azimuths equal to \{$\alpha$, $\alpha \pm 45^{\circ}$, $\alpha \pm 90^{\circ}$, $\alpha + 180^{\circ}$\}, covering 360 degrees of the human head. All images, both input and generated output, maintain a zero elevation angle, ensuring consistent horizontal viewpoints.
The generated views consist of: a reconstructed front view matching the input image; left and right profiles capturing the sides of the head; and a back view that synthesizes hair structure and color based on the frontal input and learned priors. We also generate three-quarter views (left-front and right-front) to enhance facial details in the following reconstruction stage.

To generate unseen views of the human head, we reformulate view synthesis from a single image as a conditional diffusion process. Specifically, we employ a DDPM-based diffusion model \(f_D\) to generate \(N\) distinct views, denoted \(X_0^1, X_0^2, \dots, X_0^N\), from a single front-facing image \(y\) and corresponding text embeddings \(e^1, e^2, \dots, e^N\). This process can be expressed as:
\begin{equation}
    \{X_0^1, X_0^2, \dots, X_0^N\} = f_D\Bigl(y, \{e^1, e^2, \dots, e^N\}\Bigr).
\end{equation}
Our objective is to learn the joint distribution of these views conditioned on the input image and text embeddings. We denote this joint distribution as:
\begin{equation}
    p_\theta(x_0^{1:N} \mid y, e^{1:N}) \equiv p_\theta\Bigl(\{x_0^1, \dots, x_0^N\} \mid y, \{e^1, \dots, e^N\}\Bigr).
\end{equation}
In the following discussion, we omit the condition $y$ and $e^1, e^2, \dots, e^N$ for simplicity. The joint distribution as $p_\theta(x_0^{1:N})$ is characterized by a Markov Chain (reverse process):
\begin{equation}
\begin{aligned}
    p_\theta(x_{0:T}^{1:N}) &= p_\theta(x_T^{1:N})\prod_{t=1}^T p_\theta\bigl(x_{t-1}^{1:N} \mid x_t^{1:N}\bigr) \\
    &= p_\theta(x_T^{1:N})\prod_{t=1}^T\prod_{n=1}^N p_\theta\bigl(x_{t-1}^n \mid x_t^{1:N}\bigr),
\end{aligned}
\end{equation}
where $p_\theta(x_T^{1:N}) = \mathcal{N}(x_T^{1:N}; \textsc{0}, \textsc{I})$ and $p_\theta(x_{t-1}^n\mid x_t^{1:N}) = N(x_T^n; \mu_\theta^n(x_t^{1:N}, t), \sigma_t^2\textsc{I})$. $\mu_\theta(x_t^{1:N}, t)$ is a trainable component while the variance $\sigma_t^2$ is untrained time-dependent constants. To learn $\mu_\theta$ for generation, a Markov chain called forward process is constructed as:
\begin{equation}
\begin{aligned}
    q\bigl(x_{1:T}^{1:N} \mid x_0^{1:N}\bigr) &= \prod_{t=1}^T q\bigl(x_t^{1:N} \mid x_{t-1}^{1:N}\bigr) \\
    &= \prod_{t=1}^T\prod_{n=1}^N q\bigl(x_t^n \mid x_{t-1}^n\bigr),
\end{aligned}
\end{equation}
where $q\bigl(x_t^n  \mid x_{t-1}^n \bigr) \;=\; \mathcal{N}\bigl(x_t^n ; \sqrt{1 - \beta_t}\,x_{t-1}^n ,\, \beta_t \mathbf{I}\bigr),$ and \(\beta_t\) are constants. DDPM~\cite{ddpm} shows that by defining
\begin{equation}
    \mu_\theta^n(x_t^{1:N}, t) = \frac{1}{\sqrt{\alpha_t}}\left(x_t^n-\frac{\beta_t}{\sqrt{1 -\bar{\alpha}_t}} \,\epsilon_\theta\bigl(x_t^{1:N}, t\bigr) \right).
\end{equation}
$\alpha_t$ and $\bar{\alpha}_t$ are constants derived from $\beta_t$ and $\epsilon_\theta$ is a noise predictor. We learn $\epsilon_\theta$ by 
\begin{equation}
    \ell \;=\; \mathbb{E}_{t,x_0^{1:N},n,\epsilon^{1:N}}
\bigl[\|\epsilon^n \;-\; \epsilon_\theta^n(x_t^{1:N},\,t)\|_2\bigr],
\end{equation}
where $\epsilon^{1:N}$ is the Gaussian noise of size $N\!\times\!H\!\times\!W$ added to all $N$ views, and $\epsilon_\theta^n$ is the noise predictor on the $n_{th}$ view. We provide ablation study results of the multi-view attention mechanism in Tab.~\ref{tab:rebuttal_2}.

\section{Experimental Details}
\label{sec:supp_exp_details}
\subsection{Details on Benchmark Evaluation}
\noindent\textbf{Test Camera Extrinsic.}
Both the Cafca~\cite{cafca} and Ava-256~\cite{ava256} datasets offer multi-view RGB images along with corresponding camera poses. However, their camera systems differ from those used in {\ourmethod} and the baselines.
Directly applying their camera poses for inference is infeasible. Hence, we recalculate the test camera extrinsic in each method's camera system with the following procedure.
The Ava-256 dataset uses a world coordinate system with the origin set at one of the camera positions.
We first re-center the world coordinate origin to the midpoint of all camera locations, which is approximately the center of the human head.
This step is unnecessary for the Cafca dataset, as its world coordinate origin is defined as the head's center.
Next, we compute the rotation transformation from the test camera pose to the input camera pose within the dataset’s coordinate system.
We then apply the same transformation to the input camera pose in each method’s camera system and rescale the translation to match the settings of each method to get the test camera extrinsic under each method's camera system. After applying the camera pose transformation, perfect alignment is not achieved due to differences in camera distance and intrinsic parameters. To address this, we manually crop and scale the rendered images for closer alignment with the target images.

\vspace{1mm}
\noindent\textbf{Facial Landmark Alignment.}
To align two images based on their facial landmarks, we first compute the geometric transformations—scale and translation—that align the landmarks of one image with the landmarks of the other. Given an input image \( I_1 \) and two sets of corresponding facial landmarks \( L_1 \) and \( L_2 \), we begin by calculating the centroids of the landmark sets, centering the landmarks around their respective centroids. Next, we compute the uniform scaling factor and translation vector that minimize the difference between the centered landmarks. These transformations are then applied to the input image \( I_1 \), producing the transformed image \( I_t \) in which the facial landmarks are aligned with those of \( L_2 \). This process is illustrated in Algorithm~\ref{alg:facial}. 

\begin{algorithm}[tbp]
    \caption{Image Alignment via Facial Landmarks}
    \KwIn{Image $I_1$, Landmarks $L_1$, $L_2$}
    \KwOut{Transformed image $I_t$}
    \SetKwFunction{GetTransform}{GetTransformFromLandmarks}
    \SetKwFunction{ApplyTransform}{ApplyTransformToImage}
    \SetKwFunction{TransformImage}{TransformImageWithLandmarks}
    \SetKwProg{Fn}{Function}{:}{}
    \label{alg:facial}
    \Fn{\GetTransform{$L_1$, $L_2$}}{
        Compute centroids $C_1, C_2$ of $L_1, L_2$\;
        Center landmarks: $L'_1 \gets L_1 - C_1$, $L'_2 \gets L_2 - C_2$\;
        Compute scale: $s \gets \frac{\sum (L'_1 \cdot L'_2)}{\sum (L'_1 \cdot L'_1)}$\;
        Compute translation: $t \gets C_2 - s \cdot C_1$\;
        \Return{$s, t$}\;
    }
    \Fn{\ApplyTransform{$I, s, t$}}{
        Create transformation matrix $M$\;
        Transform image: $I_t \gets \text{warpAffine}(I, M)$\;
        \Return{$I_t$}\;
    }
    \Fn{\TransformImage{$I_1, L_1, L_2$}}{
        Compute $s, t \gets \GetTransform(L_1, L_2)$\;
        Transform image: $I_t \gets \ApplyTransform(I_1, s, t)$\;
        \Return{$I_t$}\;
    }
\end{algorithm}

\subsection{Implementation Details}

\noindent\textbf{Multi-view Diffusion.}
Our multi-view diffusion model is built based on the open-source latent diffusion framework, Stable Diffusion V2-1-unCLIP model~\cite{unclip}. The model is trained on eight A100 GPUs (each with 80 GB of memory) using a batch size of 64 over 20,000 steps, with a learning rate of 1e-4. For classifier-free guidance (CFG)~\cite{cfg}, the CLIP condition was randomly omitted at a rate of 0.05 during training. During inference, we utilized the DDIM sampler~\cite{ddim} with 50 steps and a guidance scale of 3.0 to generate multi-view images. Both the input and output images have a resolution of 512$\times$512.

\vspace{1mm}
\noindent\textbf{Transformer-based Gaussian Reconstructor.}
The training of the reconstructor follows~\cite{gslrm}. During each training step, we randomly sample a set of 8 images (4 as input views and 4 as supervision views) from either 32 ambient light renderings or 25 random HDR environment light renderings. Both input and output images are rendered at a resolution of 512×512. The model is fine-tuned for 20,000 steps using eight A100 GPUs, each equipped with 40 GB of memory.

\begin{figure}[tb]
    \centering
    \includegraphics[width=.47\textwidth]{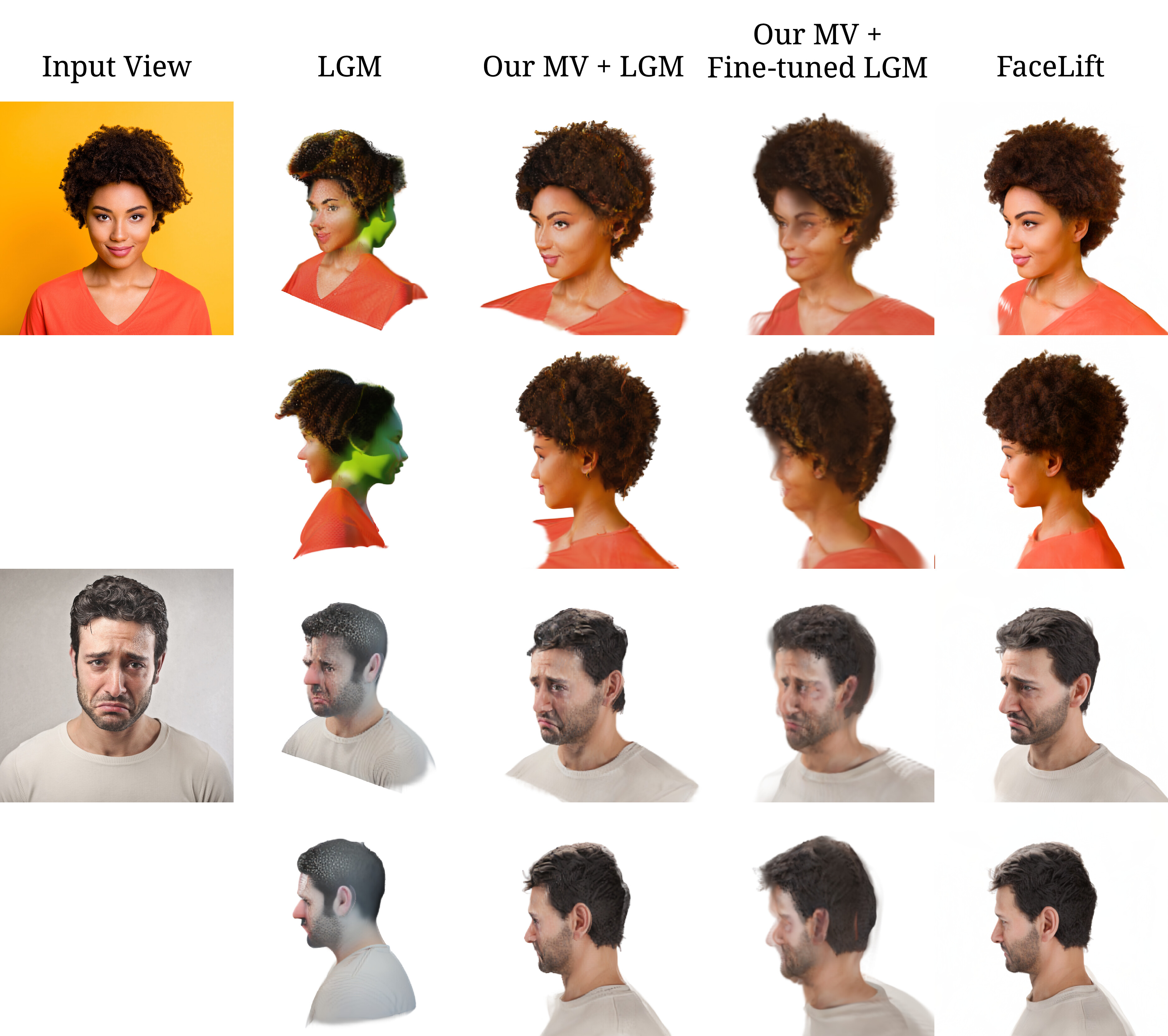}
    \caption{\textbf{Visual Comparison with LGM.} Leveraging the outputs of our multi-view diffusion model enhances the performance of LGM~\cite{lgm} (denoted as \textit{Our MV + LGM}). We further fine-tuned LGM using our synthetic human head data, resulting in \textit{Our MV + Fine-tuned LGM}; however, its performance was inferior to that achieved with the original weights in \textit{Our MV + LGM}.}
    \label{fig:supp_lgm}
\end{figure}
For a fair comparison, we also fine-tune LGM~\cite{lgm} with our synthetic data with their provided training codes. However, the fine-tuned LGM achieves inferior performance than the original weights, as shown in Fig.~\ref{fig:supp_lgm}.

\subsection{Datasets}
\noindent\textbf{Cafca Dataset.}
The Cafca dataset~\cite{cafca} comprises 1,500 identities, 30 camera poses, 13 expressions, and three environments. From this, we select 40 identities, as detailed in Tab.~\ref{tab:cafca_id}. 
We utilize the first expression and the first environment (folder 00000\_000) for each identity. The input view and test views corresponding to each identity are also specified in Tab.~\ref{tab:cafca_id}. 

\vspace{1mm}
\noindent\textbf{Ava-256 Dataset.}
The Ava-256 dataset~\cite{ava256} consists of 256 identities, each captured by 80 cameras, with over 5,000 frames per camera. For qualitative evaluation, we select 10 identities, each with 10 test camera views. All selected frames feature natural expressions. We use camera 401168 as the input view, as it captures the front view of the faces and is positioned at the center of Ava-256's world coordinate system. The input view, test view, and corresponding frame IDs are detailed in Tab.~\ref{tab:ava_256_id}.

\section{Limitations}
\label{sec:supp_limitations}
\begin{figure}[tb]
    \centering
    \includegraphics[width=.47\textwidth]{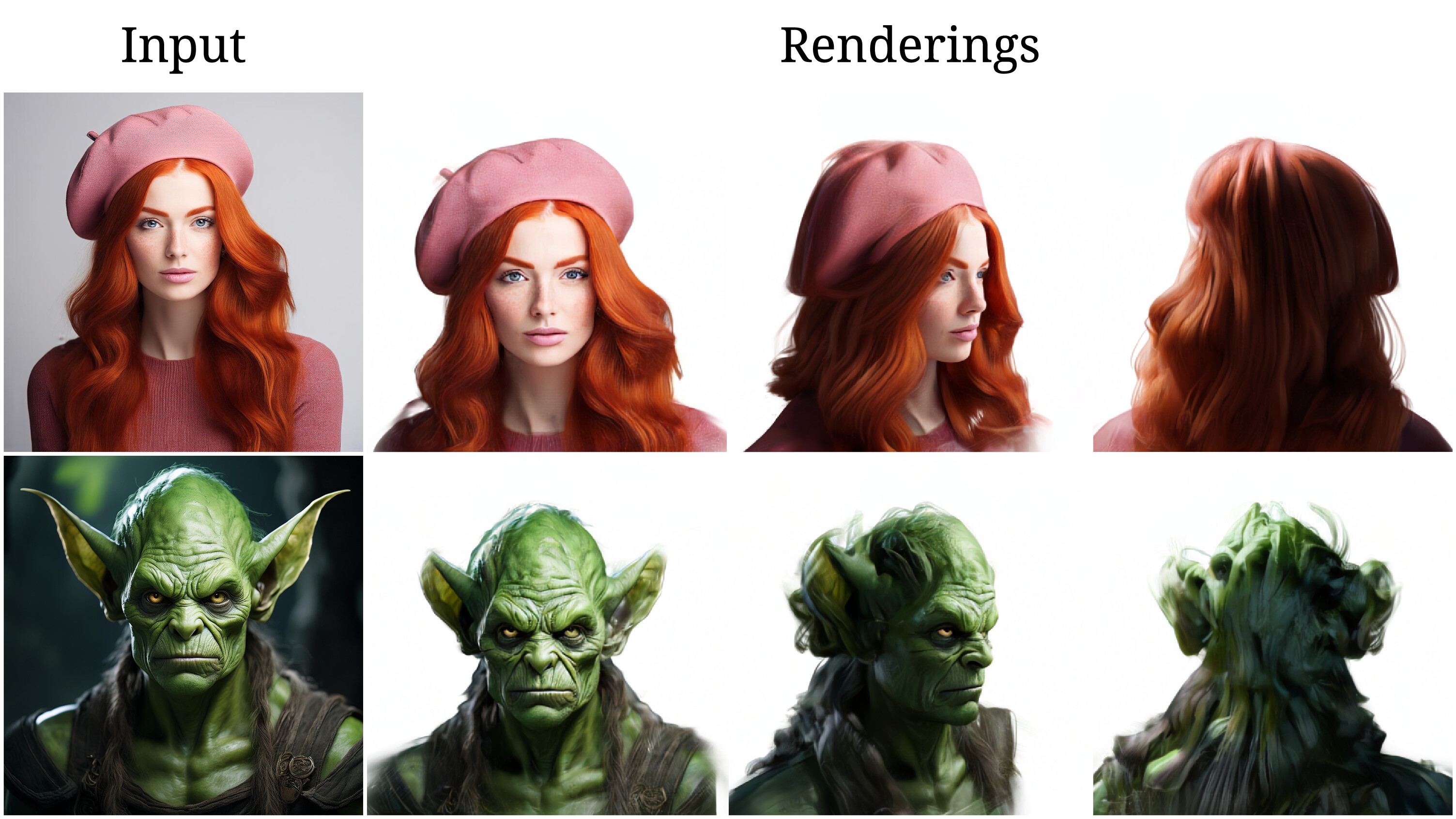}
    \caption{\textbf{Limitation of {\ourmethod}.} Due to the absence of accessories in the training data, our method often generates hair-like textures to approximate hats. Additionally, it occasionally produces extraneous hair when encountering out-of-distribution images.}
    \label{fig:supp_limitation}
\end{figure}
{\ourmethod} achieves high-fidelity, photorealistic 3D head reconstruction from a single input image. It provides detailed representations of hair and skin texture while demonstrating superior identity preservation compared to existing methods.
Despite these appealing results, our approach has certain limitations. First, our synthetic dataset does not include accessories such as hats or glasses. As a result, when the input image features a hat, the model may generate hair-like textures to approximate the back of the hat, as illustrated in Fig.~\ref{fig:supp_limitation}, row 1. This limitation could be addressed by incorporating synthetic data with accessories. Additionally, when handling out-of-distribution inputs, such as those in Fig.~\ref{fig:supp_limitation}, row 2, the model occasionally generates extraneous hair. This issue might be mitigated by refining the training data distribution or introducing text prompts to enhance control over the multi-view diffusion generation process.
Finally, in some cases, the unseen regions of the face appear more blurred than the visible areas (frontal face). Our system emphasizes detailed reconstruction of the front face: most views generated by the diffusion model concentrate on the frontal region, and the input-view reconstruction strategy faithfully preserves its features. In contrast, features of the back of the head are primarily learned from synthetic data. Additionally, when simulating lighting, the model tends to darken the back head and introduce shadows, often causing the hair to appear black.


\begin{table}[tb]
  \centering
  \scriptsize
  \begin{tabular}{@{\hskip 2.0mm}l@{\hskip 2.0mm}c@{\hskip 2.0mm}c@{\hskip 2.0mm}}
    \toprule
    ID & Input View & Test Views \\
    \midrule
    00000 & 26 & 00 02 06 08 10 11 12 13 17 19 20 23 24 26 \\
    \midrule
    00002 & 12 & 00 03 04 05 06 07 08 09 12 13 15 17 21 22 23 24 25 \\
    \midrule
    00004 & 07 & 03 04 07 09 10 11 18 19 23 24 25 26 27 29 \\
    \midrule
    00005 & 15 & 01 02 06 07 08 10 11 13 15 18 19 20 21 23 26 27 28 \\
    \midrule
    00006 & 27 & 00 02 10 19 20 23 27 \\
    \midrule
    00007 & 09 & 03 04 09 11 13 15 16 17 19 21 24 26 28 \\
    \midrule
    00010 & 24 & 02 04 08 10 12 13 14 15 17 21 22 23 24 25 26 27 28 29 \\
    \midrule
    00011 & 07 & 02 05 07 09 11 12 14 16 24 27 29 \\
    \midrule
    00014 & 03 & 02 03 06 12 14 17 22 23 25 28 29 \\
    \midrule
    00015 & 22 & 00 02 04 06 09 12 14 15 20 22 24 27 28 \\
    \midrule
    00017 & 12 & 01 02 07 12 14 15 16 17 20 22 23 24 25 26 \\
    \midrule
    00018 & 08 & 00 02 06 08 09 13 16 18 20 25 26 \\
    \midrule
    00019 & 14 & 00 04 05 06 10 12 13 14 16 17 18 20 21 22 26 28 \\
    \midrule
    00020 & 01 & 00 01 03 04 06 07 10 14 16 17 19 22 23 25 26 27 29 \\
    \midrule
    00021 & 11 & 02 03 05 07 08 09 11 14 15 17 19 21 22 23 26 \\
    \midrule
    00022 & 18 & 00 01 03 07 08 09 11 12 17 18 19 21 22 24 26 28 \\
    \midrule
    00023 & 03 & 00 03 05 06 08 12 14 18 24 25 27 \\
    \midrule
    00028 & 18 & 04 05 06 10 12 13 16 18 19 22 24 25 28 29 \\
    \midrule
    00030 & 21 & 00 01 02 03 06 07 08 11 14 17 19 21 22 24 26 \\
    \midrule
    00033 & 03 & 00 03 06 11 12 13 15 19 21 22 24 27 28 \\
    \midrule
    00034 & 10 & 01 06 07 09 10 13 15 16 17 18 19 23 25 28 \\
    \midrule
    00048 & 04 & 00 01 02 04 05 06 07 10 12 15 20 23 24 25 27 28 \\
    \midrule
    00051 & 26 & 03 07 10 11 15 17 19 21 22 24 26 28 29 \\
    \midrule
    00056 & 07 & 00 01 02 07 08 12 14 15 17 18 20 21 22 23 24 25 28 29 \\
    \midrule
    00057 & 11 & 00 01 02 03 05 06 08 11 12 14 17 18 19 22 26 29 \\
    \midrule
    00063 & 01 & 01 02 05 08 09 11 13 14 16 17 18 20 22 25 26 28 29 \\
    \midrule
    00066 & 13 & 01 05 06 07 12 13 21 22 26 27 \\
    \midrule
    00068 & 12 & 00 01 06 10 12 14 16 19 21 22 25 26 27 \\
    \midrule
    00072 & 25 & 02 04 05 10 12 13 14 17 25 26 \\
    \midrule
    00078 & 20 & 00 02 03 05 06 07 08 12 13 14 15 16 17 18 20 24 25 28 29 \\
    \midrule
    00080 & 08 & 01 03 04 05 06 08 10 12 14 15 16 17 22 24 26 \\
    \midrule
    00082 & 16 & 05 06 07 09 13 16 17 19 20 23 25 27 \\
    \midrule
    00083 & 16 & 00 02 03 04 05 08 09 13 14 16 17 19 21 22 24 25 27 29 \\
    \midrule
    00084 & 01 & 02 04 08 09 11 12 14 16 17 18 19 23 28 29 \\
    \midrule
    00086 & 13 & 00 01 03 04 08 09 13 14 17 18 19 20 22 23 24 \\
    \midrule
    00087 & 01 & 00 01 02 04 07 08 09 12 15 16 17 18 21 24 26 27 \\
    \midrule
    00094 & 08 & 02 05 08 09 12 19 24 25 27 \\
    \midrule
    00095 & 08 & 00 01 03 04 08 09 10 11 13 14 18 19 20 21 24 28 29 \\
    \midrule
    00096 & 01 & 01 05 07 10 12 17 19 21 22 28 \\
    \midrule
    00099 & 00 & 00 02 03 04 05 07 08 09 12 14 15 16 17 20 21 23 25 29 \\
    \bottomrule
  \end{tabular}
  \caption{Identities and views used for the experiment on Cafca.}
  \label{tab:cafca_id}
\end{table}

\begin{table}[tb]
  \centering
  \scriptsize
  \begin{tabular}{@{\hskip 2.5mm}l@{\hskip 2.5mm}c@{\hskip 2.5mm}c@{\hskip 2.5mm}c@{\hskip 2.5mm}}
    \toprule
    ID & Frame ID & Input View & Test Views \\
    \midrule
    \multirow{4}{*}{20210810--1306--FXN596} & \multirow{4}{*}{029693} & \multirow{4}{*}{401168} & 400944 400981 401031 \\
    & & & 401075 401163 401175 \\
    & & & 401292 401303 401316 \\
    & & & 401463 \\
    \midrule
    \multirow{4}{*}{20210827--0906--KDA058} & \multirow{4}{*}{028930} & \multirow{4}{*}{401168} & 400944 401031 401071 \\
    & & & 401163 401166 401292 \\
    & & & 401316 401408 401410 \\
    & & & 401458 \\
    \midrule
    \multirow{4}{*}{20210901--0833--LAS440} &
    \multirow{4}{*}{027655} & \multirow{4}{*}{401168} & 400944 401031 401161 \\
    & & & 401163 401172 401292 \\
    & & & 401303 401316 401410 \\
    & & & 401458 \\
    \midrule
    \multirow{4}{*}{20210929--0827--MCR809} &
    \multirow{4}{*}{029457} & \multirow{4}{*}{401168} & 400981 401070 401158 \\
    & & & 401166 401173 401305 \\
    & & & 401313 401408 401410 \\
    & & & 401458 \\
    \midrule
    \multirow{4}{*}{20211001--0855--KJJ701} &
    \multirow{4}{*}{032309} & \multirow{4}{*}{401168} & 400939 401031 401163 \\
    & & & 401166 401292 401316 \\
    & & & 401408 401410 401452 \\
    & & & 401458 \\
    \midrule
    \multirow{4}{*}{20220215--0801--ONK705} &
    \multirow{4}{*}{027201} & \multirow{4}{*}{401168} & 400944 401031 401045 \\
    & & & 401163 401166 401172 \\
    & & & 401408 401410 401463 \\
    & & & 401469 \\
    \midrule
    \multirow{4}{*}{20220310--1128--ZSC414} & \multirow{4}{*}{028601} & \multirow{4}{*}{401168} & 400942 401031 401045 \\
    & & & 401163 401164 401166 \\
    & & & 401303 401408 401410 \\
    & & & 401411 \\
    \midrule
    \multirow{4}{*}{20220712--1040--JEH262} & \multirow{4}{*}{030060} & \multirow{4}{*}{401168} &
    400944 400981 401031 \\
    & & & 401045 401163 401408 \\
    & & & 401410 401452 401458 \\
    & & & 401469 \\
    \midrule
    \multirow{4}{*}{20220809--1321--UTC375} & \multirow{4}{*}{027432} & \multirow{4}{*}{401168} & 401031 401071 401163 \\
    & & & 401166 401175 401292 \\
    & & & 401303 401452 401458 \\
    & & & 401469 \\
    \midrule
    \multirow{4}{*}{20220818--1653--SSF476} & \multirow{4}{*}{036588} & \multirow{4}{*}{401168} & 400981 401031 401071 \\
    & & & 401163 401166 401175 \\
    & & & 401408 401410 401458 \\
    & & & 401469 \\
  \bottomrule
  \end{tabular}
  \caption{Identities and views used for the experiments on Ava-256.}
  \label{tab:ava_256_id}
\end{table}

\end{document}